# Journey into Automation: Image-Derived Pavement Texture Extraction and Evaluation

Bingjie Lu[1], Han-Cheng Dan[2*], Yichen Zhang[3], Zhetao Huang[4]

**Bingjie Lu,** postgraduate, School of Civil Engineering, Central South University, Hunan, Changsha 410075, China, Email: bingjie_lu@csu.edu.cn

**Han-Cheng Dan**, professor, Ph.D., School of Civil Engineering, Central South University, Hunan, Changsha 410075, China; National Engineering Research Center for High-speed Railway Construction Technology, Central South University, Changsha, Hunan 410075, China; Rail Data Research and Application Key Laboratory of Hunan Province, Central South University, Changsha, Hunan 410075, China, Email: danhancheng@csu.edu.cn

**Yichen Zhang,** postgraduate, School of Civil Engineering, Central South University, Hunan, Changsha 410075, China, Email: yichen_zhang@csu.edu.cn

**Zhetao Huang,** postgraduate, School of Civil Engineering, Central South University, Hunan, Changsha 410075, China, Email: zhetao_huang@csu.edu.cn

*Corresponding author: Han-Cheng Dan, Email: danhancheng@csu.edu.cn

**Abstract:** Mean texture depth (MTD) is pivotal in assessing the skid resistance of asphalt pavements and ensuring road safety. This study focuses on developing an automated system for extracting texture features and evaluating MTD based on pavement images. The contributions of this work are threefold: firstly, it proposes an economical method to acquire three-dimensional (3D) pavement texture data; secondly, it enhances 3D image processing techniques and formulates features that represent various aspects of texture; thirdly, it establishes multivariate prediction models that link these features with MTD values. Validation results demonstrate that the Gradient Boosting Tree (GBT) model achieves remarkable prediction stability and accuracy ($R^2$ = 0.9858), and field tests indicate the superiority of the proposed method over other techniques, with relative errors below 10%. This method offers a comprehensive end-to-end solution for pavement quality evaluation, from images input to MTD predictions output.

**Keywords:** Pavement engineering; Mean texture depth; 3D reconstruction; Image processing; Multivariate regression.

# 1. Introduction

The asphalt pavement texture plays a crucial role in determining friction, splash and spray, and rolling resistance. Furthermore, it serves as a means to identify segregation or non-uniformity and to assess the noise characteristics of pavements [1]. Additionally, the evaluation of pavement wear and the design of surface texture rely heavily on texture data [2]. Accordingly, it is important to monitor pavement texture periodically and evaluate the safety performance of asphalt pavement.

The mean texture depth (MTD) and the mean profile depth (MPD) are the most commonly utilized parameters for characterizing pavement texture [3]. However, the sand patch method (SPM), which is extensively used in detection systems to calculate the MTD, is marred by significant drawbacks, such as low efficiency and the potential to disrupt traffic flow [4]. Moreover, the MPD data derived from scanning fail to adequately represent the relationship between pavement texture and its performance, contrary to expectations [5].

Recent studies have focused on evaluating pavement characteristics using 3D data, demonstrating that 3D texture measurement offers a more precise reflection of pavement's physical attributes [6]. Laser technology, in particular, facilitates the acquisition of accurate 3D texture data, such as point clouds and depth maps, through a non-contact approach, enhancing the precision of MTD computation [7]. However, the prohibitive cost of laser equipment and the complexity of 3D data post-processing, which demands specific technical skills from operators, limit the method's widespread application. An alternative, the 3D Image-Based Texture Analysis Method (3D-ITAM), acquires 3D data from a single digital image [8]. Yet, a single image often fails to capture texture details comprehensively, and the resultant 8-bit precision grayscale image lacks the detail for extracting precise texture features, leading to suboptimal MTD prediction accuracy. Additionally, the extraction of features for MTD prediction often involves switching between multiple image analysis software tools. The rationale behind the need to extract certain features is also challenging to explain, and the prediction of MTDs often relies on fitting empirical models [9], which are only applicable to specific datasets and cannot be generalized to other situations. The Structure from Motion (SfM) technique, while achieving commendable 3D reconstruction results from multi-view digital images, requires skilled personnel and non-open-source commercial software to operate [10]. Additionally, the digital camera utilized, being monocular, cannot deliver depth information at real-world scales comparable to that of expensive laser devices or binocular cameras, which are capable of providing absolute depth values for point cloud or depth map data [11, 12]. This limitation necessitates manual calibration prior to MTD prediction, thus impeding full automation unless custom capture devices with preset shooting angles and heights are used [13]. MTD prediction through depth maps generated from multi-view digital images via an open-source deep learning model, trained on extensive pavement data, has shown higher accuracy than SfM techniques, which are better suited for large-scale object reconstruction rather than pavements [14]. Still, the use of an inexpensive monocular camera requires manual calibration to convert relative depth information in the depth maps into absolute depth values, preventing full automation. While binocular reconstruction techniques automate stereo matching of corresponding pixels in left and right images using triangulation principles [15], the accuracy of pixel-based stereo matching is compromised by asphalt pavements' dark color and weak regional texture, failing to meet expected MTD measurement precision [16]. Table 1 summarizes representative studies of the aforementioned methods, indicating that, despite extensive research into 3D data acquisition methods, no single approach has yet achieved an economical, convenient, highly accurate, and automated measurement of MTD.

Table 1. Summary of MTD calculation methods.

| Detection methods | MTD calculation methods | MTD measurement accuracy | Comments |
|---|---|---|---|
| Laser scanning technology with a laser projector and a camera [17]. | Extract the laser data, process the 3D point cloud, and | Mean absolute error = 0.017 mm, Pearson correlation | High precision, yet the devices are costly and data |

| | obtain subblock size to evaluate MTD. | coefficient = 0.9864. | extraction is relatively complex. |
|---|---|---|---|
| 3D-ITAM with a monocular camera [8]. | Capture an image, grayscale it, and predict MTD based on a relationship model. | $R^2$ = 0.8745 | Economical and convenient, but not highly accurate. |
| SfM technique with a monocular camera [13]. | Reconstruct point clouds from multi-view images, and use mathematical models for MTD prediction. | $R^2$ = 0.9275, average deviations = 0.005 mm. | High accuracy, but requires commercial software and manual calibration. |
| Deep learning method with a monocular camera [14]. | Infer depth maps from multi-view images to calculate MTD. | $R^2$ = 0.9807, relative error = 11.72% | High accuracy, but necessitates manual calibration. |
| Binocular reconstruction with two cameras and multi-line laser generators [16]. | Reconstruct a 3D model, process and match images, and estimate MTD based on calculated MPD. | The $R^2$ values for all four test pavements are below 0.90. | Unable to achieve high precision due to pavement characteristics. |

In response to the aforementioned issues, this paper introduces an image-based, high-accuracy, calibration-free, end-to-end system that does not require expensive sensors or customized instruments. Based on the algorithm and model developed in this paper, users only need to use a digital camera to capture multi-view images of the pavement, enabling convenient MTD prediction. The main contributions of this study are summarized as follows:

- Utilizing a monocular digital camera and computer, an image-based 3D reconstruction deep learning model was employed to acquire 32-bit high-precision 3D relative depth maps of pavement textures.
- Improved 3D depth map processing methods were designed to eliminate noise and mitigate the effects of camera poses, thus enhancing the precision of texture information extraction.
- Features characterizing pavement texture were extracted based on relative depth information from 3D data to predict MTD, thus avoiding manual calibration. The analysis of the relationship between features and MTD demonstrated how each feature affects the texture depth from distinct perspectives.
- Multivariate regression models were established for MTD prediction, and the optimal prediction model was selected through performance evaluation. Field tests were conducted to verify the accuracy and superiority of the proposed method compared to other techniques.

The remainder of this paper is organized as follows. Section 2 includes the methods for texture reconstruction, 3D data processing, texture feature extraction, feature analysis, and MTD prediction model establishment. The results are presented in Section 3, respectively. The conclusion of this study is offered in Section 4.

## 2. Methodological framework

As illustrated in Fig. 1, a system for automatic MTD measurement has been developed, requiring only a digital camera and a computer. Supplemental lighting is necessary only when ambient light is insufficient. The use of affordable and readily available devices makes it convenient for users to develop their own hardware-software interaction platform. The specific operation of the system is highlighted in the blue box in Fig. 2. The algorithms and models (indicated by the green box) are developed in a Python 3.7 environment, running on a Windows 10 operating system with an AMD Ryzen 9 5900HX CPU, enabling end-to-end image input and prediction results output. Specific images or data can be obtained at each step (indicated by the pink box), facilitating the development of data display interfaces and management systems (indicated by the purple box).

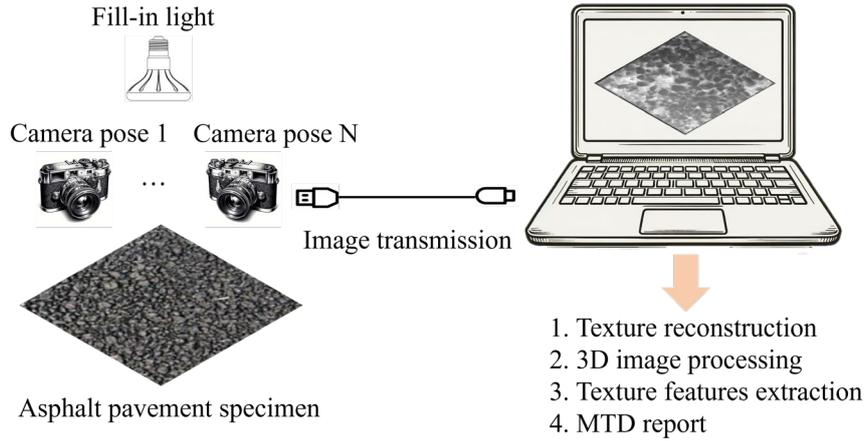

Fig. 1. The proposed automatic measurement system.

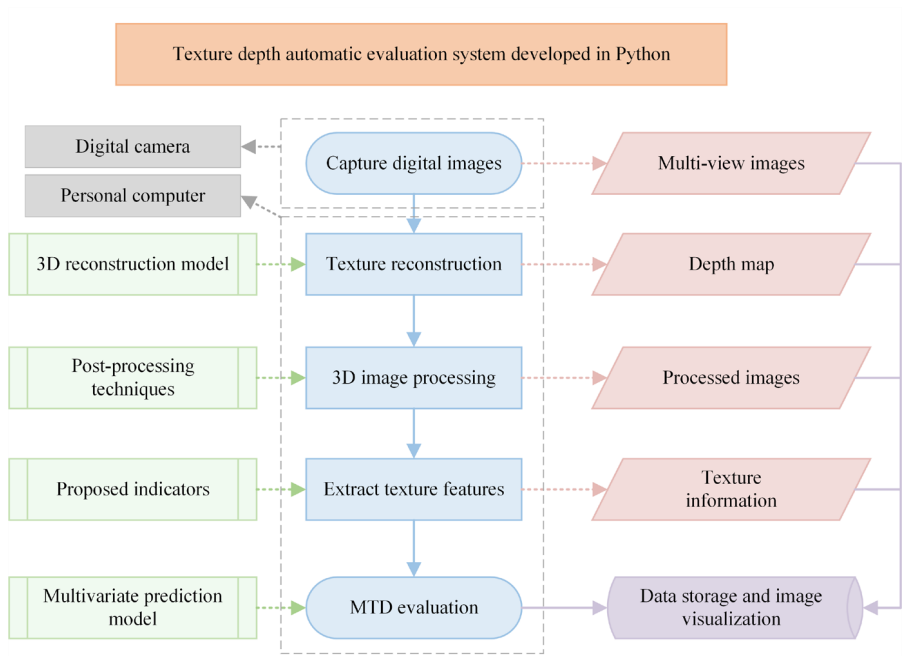

Fig. 2. The operating principle of the measurement system.

## 2.1. Pavement texture reconstruction techniques

High-precision pavement textures can be reconstructed using a deep learning model called Patchmatchnet, with further technical details available in literature [14]. This section will provide a brief introduction to this reconstruction method.

### 2.1.1. Depth map inference: Principles and applications

The model Patchmatchnet is fundamentally based on the principle of Multi-View Stereo (MVS) to infer depth maps [18]. As illustrated in Fig. 3, it defaults to selecting the first image from the input set as the reference image, with the subsequent images serving as source images to deduce depth information for the reference image. Therefore, to obtain depth information of the texture, multi-angle images of the pavement need to be captured, while ensuring that the optical axis of the first image is perpendicular to the surface. The prediction of MTD is based on the generated reference depth map.

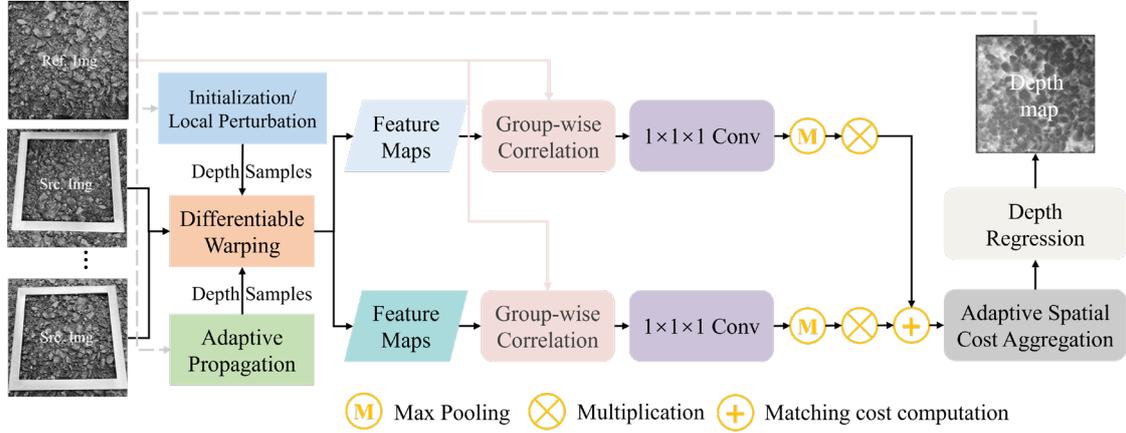

Fig. 3. The principle of depth map inference.

*2.1.2. Digital image capture method*

As depicted in Fig. 4(a), each set of multi-angle image collections comprises one reference image and 12 source images, a decision based on prior experimentation [14]. Fig. 4(b) showcases the use of a homemade steel hollow square board, featuring an inner side length of 10 cm. Digital images were captured using the built-in camera of a smartphone. The photographs were taken in a nine-grid shooting mode, aligning the board's inner edges to standardize the shooting height at 15 cm and secure consistent pixel sizes for the reference images. While this study did not enforce strict distance criteria for the source images, for optimal stereo matching accuracy [19-21], they were captured from an approximate distance of 20 cm from the pavement, close to the distance from the reference image to the surface. Given that the core of reconstruction hinges on feature matching across a series of images [22, 23], a shooting angle ranging between 30° and 40° among source images is recommended to ensure a high degree of image overlap [24, 25]. As the primary objective is to retrieve depth information for the reference image, it's vital that the overlap between source and reference views is at least 70% [26, 27]. The resolution of the collected images is 3024×3024, a quality that was confirmed to accurately reconstruct textures [14]. Furthermore, to maintain uniform lighting conditions, photos were taken at the same time of day.

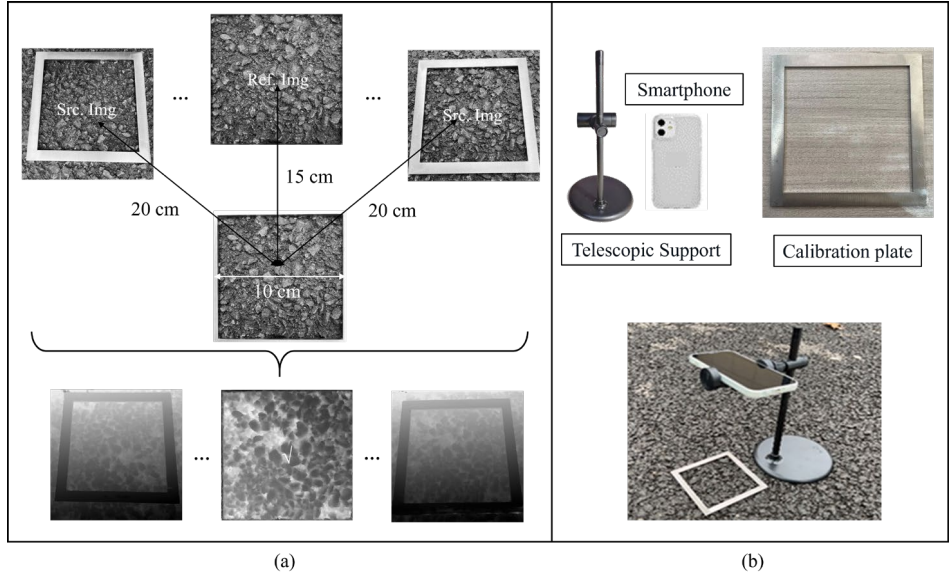

Fig. 4. (a) Requirements for shooting distance and angles; (b) Shooting devices.

*2.2. 3D data processing strategies*

To enhance the accuracy of texture information extraction in subsequent Section 2.3, a 3D image processing workflow is designed. This workflow includes data normalization, data smoothing, as well as tilt and offset calibration.

*2.2.1. Depth value normalization*

Due to the changing camera relative poses when capture devices without strict fixed shooting angles and distances are used, the scale of depth values in reference depth maps may vary [28]. This variation can result in texture feature information extracted from depth maps being incompatible with the same prediction model [29]. An effective solution to this issue is the application of the max-min normalization algorithm [30]. The normalized *z*-axis value is derived by:

$$z_{norm} = \frac{z - z_{min}}{z_{max} - z_{min}} \tag{1}$$

where *z* and $z_{norm}$ indicate the relative depth and normalized depth values, respectively. $z_{max}$ and $z_{min}$ are the maximal and minimal values of the depth data. This formula ensures that all depth data are scaled to a common range, making the texture information extracted from them uniform across different captures [31, 32].

*2.2.2. Data smoothing*

Depth maps derived from multi-view reconstruction can be compromised by unnecessary noise, which may stem from issues such as image mismatches, significant variances in brightness and contrast between images, and limitations inherent to the depth estimation algorithms [18]. Widely used filtering algorithms, like the median filter and the mean filter, show effectiveness for 2D images [33-35]. However, they fail to preserve the fine details in 3D data [36]. The bilateral filter has been demonstrated to be effective for preserving information in 3D images [37] and serves as the primary inspiration for the local adaptive filtering algorithm developed in this study. For each pixel, this algorithm computes the local mean depth and the depth variance within its surrounding neighborhood. Then, the depth values are adjusted based on the ratio of the noise variance (estimated from the original image's noise) to the local variance:

$$z' = z - \frac{\sigma_\eta}{\sigma_{S_{xy}}} \cdot (z - z_{S_{xy}}) \qquad (2)$$

where $z'$ is the denoised depth value, $z$ is the depth value in the noisy image, $\sigma_\eta$ is the noise variance of the original image, $\sigma_{S_{xy}}$ is the depth value variance in the local neighborhood, and $z_{S_{xy}}$ represents the average depth value in the local neighborhood.

*2.2.3. Tilt & offset correction*

Due to the presence of pavement slopes (as shown in Fig. 5(b)) and potential tilts between the shooting angle and the pavement (as depicted in Fig. 5(c)), reference depth maps may not reflect the depth information between pixels as accurately as in the ideal scenario illustrated in Fig. 5(a). To address this issue, two methods are proposed.

The first method involves utilizing the Random Sample Consensus (RANSAC) algorithm [38], renowned for its effectiveness in fitting planes to 3D data, to correct errors. The equations are as follows:

$$z = a \cdot x + b \cdot y + c \tag{3}$$

$$z'' = z' - z \tag{4}$$

where $x$ and $y$ are the horizontal and vertical coordinates in the depth map, respectively, $a$, $b$, and $c$ are the parameters calculated through the plane fitting process. $z$ is the plane fitting value. $z'$ and $z''$ denote the depth data before and after correction, respectively.

The second method applies the RANSAC algorithm to derive a polynomial for surface fitting [39, 40]. A third-degree polynomial was selected for its versatility in adapting to complex surfaces, effectively balancing the need for accuracy with the risk of overfitting, which is more prevalent with higher-degree polynomials:

$$z = a + bx + cy + dx^2 + exy + fy^2 + gx^3 + hx^2y + ixy^2 + jy^3 \tag{5}$$

where $a$, $b$, $c$, $d$, $e$, $f$, $g$, $h$, $i$, and $j$ are the coefficients obtained from the fitting process. Subsequently, the surface fitting values $z$ are subtracted from the original data using Equation (4).

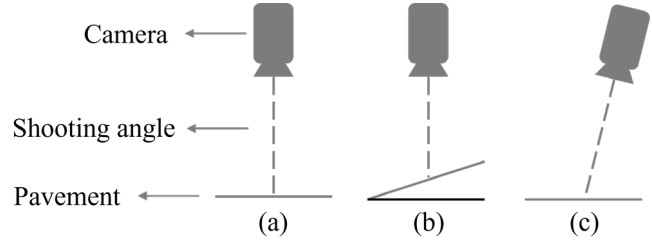

Fig. 5. (a) Ideal shooting condition; (b) Pavement with slope; (c) Tilted shooting angle.

## 2.3. Texture feature extraction approaches

This section introduces features extracted from the relative depth map, each reflecting specific aspects that influence texture depth. The extraction process lays the foundation for Section 2.6, where models are established to understand the relationship between the feature variables (texture features) and the target variable (MTD), enabling automated MTD prediction without the need for manual selection of calibration points to recover absolute depth information [14].

### 2.3.1. Relative concavity ratio

When employing the sand patch method for MTD measurement, the convex portions of the texture remain uncovered by sand after evenly spreading it, whereas the concave portions are filled. The depth and size of the concavities correlate directly with the MTD. Inspired by this principle, the proposed feature is designed to identify relatively concave parts within the texture region. To accomplish this, depth maps undergo threshold segmentation to distinguish concave from convex parts.

Fig. 6 illustrates the segmentation results using four distinct depth value thresholds. Intersection over Union (IoU) serves as the evaluation metric for threshold segmentation, where processed images are compared with manually annotated labels derived from sand patch method outcomes. IoU is calculated as the intersection of the manually labelled convex parts and those identified through image segmentation (the black parts), divided by the union of the total number of black pixels identified by both methods [41]. An ideal IoU score is 100%, with higher values indicating more accurate threshold selection [42, 43]. Table 2 presents the segmentation results for various mixtures across different thresholds, including calculated averages.

According to Fig. 6, segmentation under thresholds of 0.33-0.35 closely matches the labels, with minimal discrepancies, suggesting these values are near the optimal threshold. Table 2 indicates that the average segmentation result across four mixtures is highest at 86.34% under a 0.35 threshold, with individual results all exceeding 80%. Segmentation effectiveness declines when the threshold deviates from 0.35.

Since threshold determination relies on the map's depth information, calculating the area proportion of concave parts from the segmented image effectively reflects their volumetric proportion in three dimensions. The concavity proportion, indicated by *P*, can be calculated as follows:

$$P = \frac{N_C}{W \cdot H} \tag{6}$$

where $N_c$ represents the number of pixels in the concave part, and *W* and *H* represent the total number of pixels in the *x* and *y* directions, respectively.

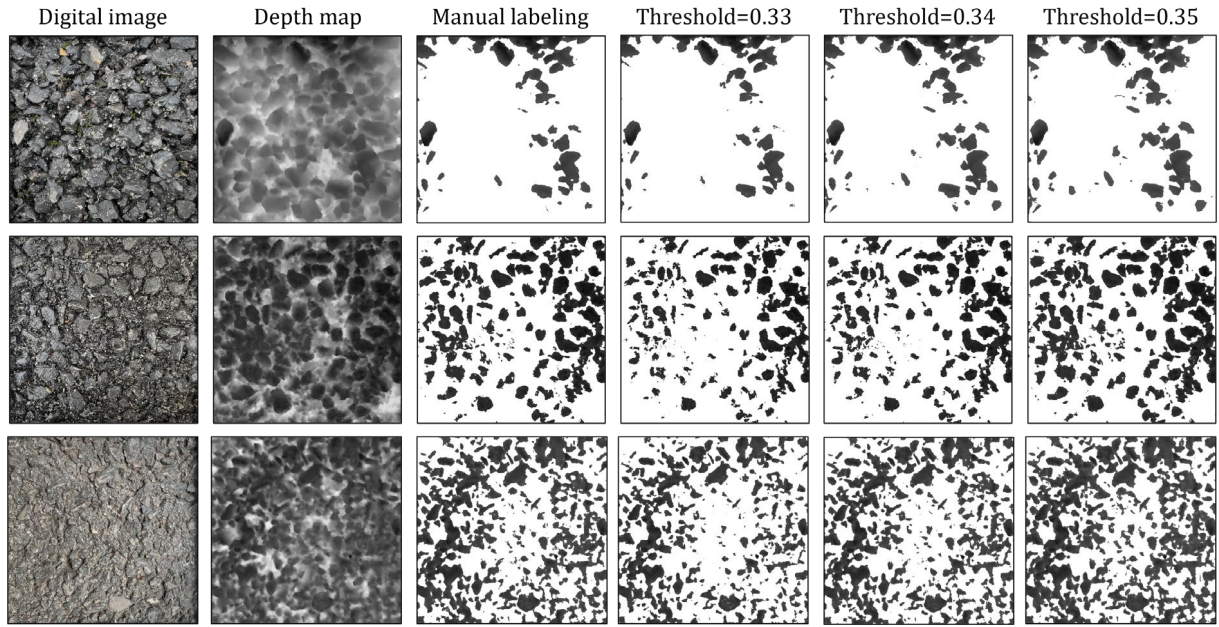

Fig. 6. The segmentation results obtained using the four thresholds.

Table 2. IoU results of the four thresholds.

| Threshold | AC-13 | AC-16 | SMA-13 | OGFC-16 | Average IoU (%) |
|---|---|---|---|---|---|
| 0.33 | 77.38 | 69.64 | 75.42 | 71.37 | 73.46 |
| 0.34 | 84.90 | 80.25 | 81.93 | 78.54 | 81.41 |
| **0.35** | 81.86 | 88.37 | 90.15 | 84.96 | **86.34** |
| 0.36 | 70.12 | 78.40 | 80.77 | 76.49 | 76.45 |

*2.3.2. Maximum particle size*

As the proportion of coarse aggregates increases, texture depth and pavement skid resistance are both enhanced [44]. The size of larger particles is a critical factor in assessing texture depth. While aggregate sizes are specified in the gradation chart (JTG E42-2005, Test Methods of Aggregate for Highway Engineering), achieving full automation in the MTD evaluation and facilitating detection where gradation information is absent require the application of image processing methods to acquire aggregate size information. A workflow for extracting aggregate size data is proposed and depicted in Fig. 7.

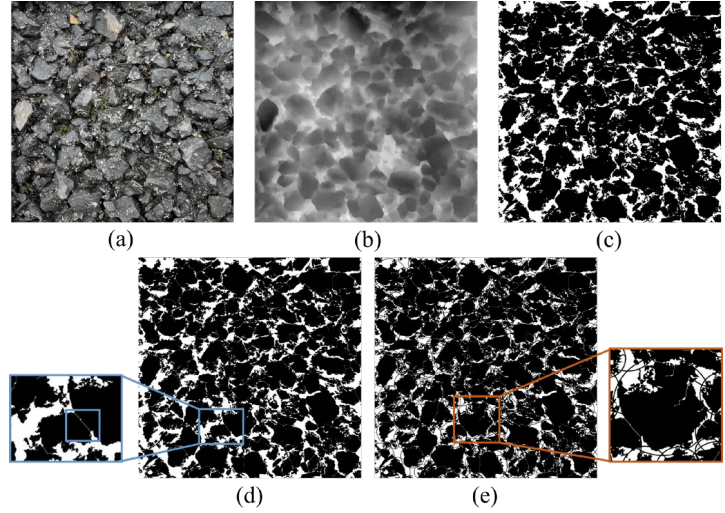

Fig. 7. The process of aggregate size data extraction: (a) The digital image; (b) The depth map; (c) Separating the aggregates from the background; (d) Separating the agglomerated particles; (e) Extracting particle size information.

(1) Binarization

To extract the size information of particles, a binarization process was implemented to segment the particles from the background. The result is illustrated in Fig. 7(c). This process involves comparing the depth values of the pixels to a threshold value, setting the particles to black and the background to white [45]. Given the necessity to separate each particle from its immediate background, the binarization algorithm must dynamically determine thresholds based on the local characteristics of the map [46], ensuring robustness to variations in depth values across different areas [47]. Consequently, a local adaptive thresholding method has been employed. The local threshold $T(x, y)$ can be determined as [48]:

$$T(x, y) = m(x, y)[1 + k(\frac{\partial(x, y)}{1 - \partial(x, y)} - 1)] \qquad (7)$$

where $m(x, y)$ is the local arithmetic mean at $(x, y)$, representing the average of the depth values within a $w \times w$ window of the depth map $I$; $\partial(x, y) = I(x, y) - m(x, y)$ is the local mean deviation, and $k$ is a bias factor that controls the level of threshold adaptation. The range of $k$ is limited to [0,1].

(2) Adhesion segmentation

During the paving process of asphalt pavements, particles within the mixtures tend to adhere to one another, especially between small and large particles [49]. This adhesion results in similar depth values for the adhered particles, making it challenging for binarization algorithms based on local thresholds to separate them. To address this issue, the adjustable watershed algorithm based on the Euclidean Distance Map, which has been proven effective in separating adhesive particles, is employed for segmentation [50]. The segmentation result is depicted in Fig. 7(d). This step is crucial to avoid errors in determining particle size in the subsequent step. Agglomerated particles can lead to an overestimation of the maximum particle size extracted, particularly when two large particles are stuck together.

(3) Particle size extraction

Various methods have been explored to determine the size of irregularly shaped particles, focusing on either theoretical significance or the practical ease of measurement. Among these methods, Feret's Diameter, defined as the longest distance between any two points along the boundary of a particle, is widely used in civil engineering for microscopic quantitative analysis [51]. Inspired by this concept, as illustrated in Fig. 7(e), all particles are enclosed by circumscribed circles, allowing for the calculation of the maximum particle size as follows:

$$D = \frac{D_{max}}{W} \tag{8}$$

where $D_{max}$ is the number of pixels of the maximum circumscribed circle's diameter, and $W$ is the total number of pixels in the *x*-direction of the image.

### 2.3.3. Aggregate voids

Considering that macrotexture is influenced by the spacing and arrangement of aggregates, particle voids can be extracted to characterize texture depth [52]. The void is defined as the largest space between a particle and all surrounding particles adjacent to it. With the Fig. 7(e) obtained from the previous step, the voids are calculated by subtracting the sizes of the particles from the distance between the centers of two adjacent circumscribed circles. This is essentially done by subtracting the number of black pixels within the arrow's span from the total arrow pixel count, as shown in Fig. 8. The aggregate void of the entire texture region can be calculated as follows:

$$K = \frac{\sum_{i=1}^{N} K_{i\_max}}{W} \tag{9}$$

where $W$ is the total number of pixels in the $x$-direction of the image, $N$ is the total number of particles, and $K_{i\_max}$ is the result of aggregate void calculation for the $i_{th}$ particle.

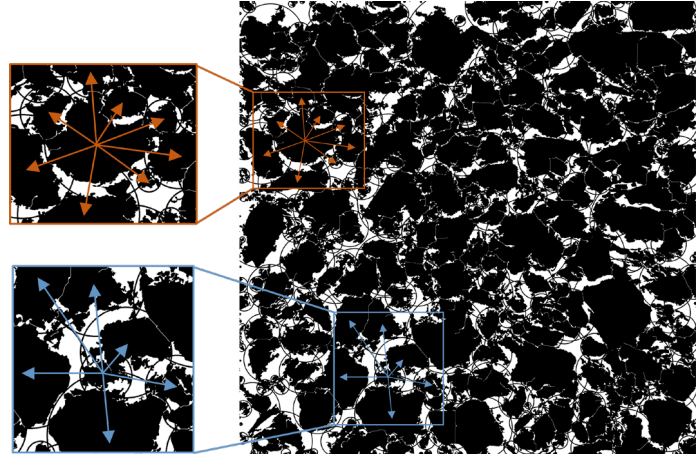

Fig. 8. $K$-feature computation schematic.

*2.4. A Feature-Label Dataset: Foundations for predictive modeling*

As illustrated in Fig. 9, the digital image collection encompassed four different types of mixtures: asphalt concrete AC-13, asphalt concrete AC-16, stone mastic asphalt SMA-13, and open graded friction course OGFC-16, culminating in a total of 160 datasets, with each type contributing 40 sets. Concurrently, the sand patch test was utilized to acquire MTD label values at each measurement point. Subsequently, texture reconstruction, image processing, and texture feature extraction were performed on 160 sets of data to construct a Feature-Label dataset. Table 3 provides partial calculation results from the dataset.

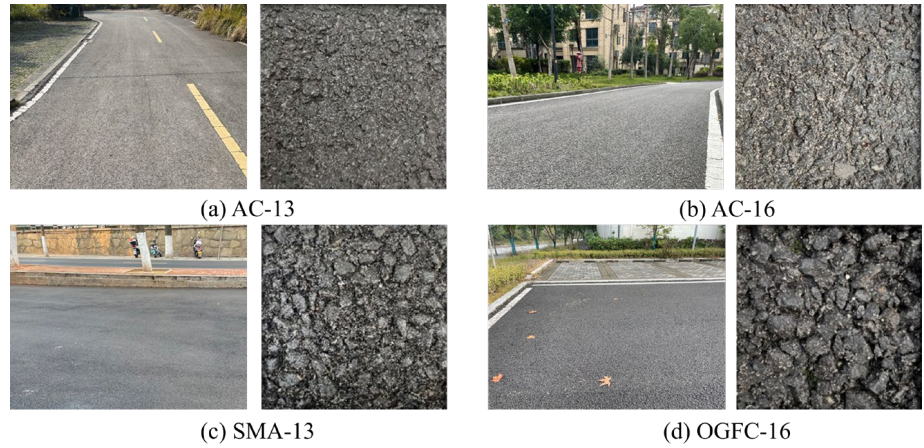

(a) AC-13          (b) AC-16

(c) SMA-13       (d) OGFC-16

Fig. 9. The field data collection.

Table 3. Partial calculation results of label and feature values.

| Identifier | MTD (mm) | $P$ | $D$ | $K$ (%) |
| --- | --- | --- | --- | --- |
| AC-13-1 | 0.57 | 0.43 | 0.14 | 1.21 |
| AC-13-2 | 0.59 | 0.48 | 0.13 | 1.18 |
| AC-13-3 | 0.62 | 0.51 | 0.15 | 1.39 |
| AC-16-10 | 0.90 | 0.70 | 0.18 | 2.19 |
| AC-16-11 | 0.93 | 0.73 | 0.20 | 2.13 |
| AC-16-12 | 0.96 | 0.77 | 0.19 | 2.24 |
| SMA-13-5 | 0.63 | 0.50 | 0.13 | 1.84 |
| SMA-13-6 | 0.67 | 0.56 | 0.15 | 1.97 |
| SMA-13-7 | 0.69 | 0.59 | 0.14 | 1.90 |
| OGFC-16-35 | 2.16 | 0.82 | 0.19 | 2.90 |
| OGFC-16-36 | 2.26 | 0.84 | 0.19 | 3.06 |
| OGFC-16-37 | 2.37 | 0.87 | 0.20 | 3.23 |

*2.5. Analytical techniques for texture feature evaluation*

To further elucidate the relationship between the proposed features and texture depth, 70% of the Feature-Label dataset was utilized to establish univariate and multivariate linear regression models, as shown in Table 4. The remaining 30% of the dataset was used for model evaluation [53, 54]. Evaluation metrics include the Sum of Squared Errors (*SSE*), the Adjusted $R^2$ to penalize model complexity, and the p-value to assess the significance probability:

$$SSE = \sum_{i=1}^{n}(y_i - \hat{y}_i)^2 \tag{10}$$

$$Adjusted\ R^2 = 1 - \frac{(1-R^2)\times(n-1)}{n-m-1} \tag{11}$$

where *n* is the number of observations, $y_i$ is the label value for the $i_{th}$ observation, $\hat{y}_i$ is the predicted values, and *m* is the number of features. When the *SSE* approaches 0, the Adjusted $R^2$ approaches 1, and the p-value decreases, it indicates that the feature variables in the linear regression model have a more significant impact on the dependent variable [55-57].

Table 4. The established linear regression models.

| Model | Feature Variables | Target Variable |
| --- | --- | --- |
| Univariate linear regression model | P/K/D | MTD |
| Multivariate linear regression model | P+K/ P+D/ K+D/ P+D+K | |

A multicollinearity analysis among the features was also conducted. Multicollinearity refers to the high linear correlation among feature variables in a regression model [58, 59], which can lead to inaccurate estimates of labels and diminish the explanatory power of multivariate models [60, 61]. Multicollinearity was diagnosed through correlation analysis and the calculation of Variance Inflation Factors (*VIF*) as follows:

$$R^2(y,\hat{y}) = 1 - \frac{\sum_{i=1}^{n}(y_i-\hat{y}_i)^2}{\sum_{i=1}^{n}(y_i-\bar{y})^2} \tag{12}$$

$$VIT = \frac{1}{1-R^2} \tag{13}$$

where $R^2$ is the coefficient of determination with one feature variable serving as the dependent variable and the others as independent variables. If the coefficient of determination exceeds 0.8 and *VIF* surpasses 10, it indicates a potential presence of multicollinearity [62, 63].

*2.6. Development of multi-feature predictive models*

Based on the feature analysis results in Section 3.3, it is advantageous for the prediction model to effectively handle correlations among variables. To this end, four machine learning algorithms were chosen to develop multi-feature prediction models: Random Forest (RF) [64], Multi-layer Perceptron (MLP) [65], Support Vector Machines (SVM) [66], and Gradient Boosting Trees (GBT) [67]. These models are well-regarded for their capacity to manage feature correlations and are known to excel in prediction accuracy and robustness [68-70]. Table 5 displays the input combinations in multivariate prediction models. Among them, RF constructs multiple decision trees, each being a nonlinear model capable of freely capturing complex relationships between features [71-73]. It makes regression predictions by combining the results of individual trees (Fig. 10(a)). MLP features a multi-layer neural network structure [74, 75], which learns the complex relationships between features and labels by combining feature values in hidden layers through both linear and nonlinear connections [76, 77], thus accomplishing prediction tasks (Fig. 10(b)). SVM achieves predictions by finding a hyperfunction $f(x) = w^T x + b$ that best fits data points [78, 79]. As illustrated in Fig. 10(c), values within the error $\varepsilon$ margin on either side of the function can be considered correctly predicted, while values outside the dashed lines require loss computation [80, 81]. The model is optimized by minimizing total loss. GBT builds a stronger predictive model by integrating multiple regression trees [82]. During training, each weak learner learns from the predecessor, cumulating in a final prediction that integrates all learners' predictions (Fig. 10(d)). By incorporating algorithms grounded in varied fundamental principles, the analysis is equipped to uncover a wider spectrum of patterns and interrelations within the data [83]. Fig. 11 illustrates the data allocation for the training, validation, and testing phases of the multivariate prediction models.

Table 5. The established multivariate regression models.

| Model | Feature Variables | Target Variable |
|---|---|---|
| RF/MLP/SVM/GBT | P+K/ P+D/ K+D/ P+D+K | MTD |

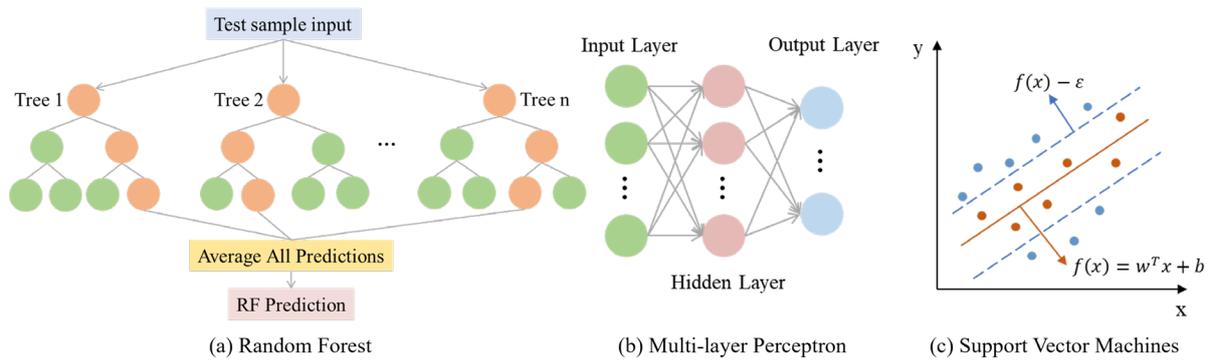

(a) Random Forest    (b) Multi-layer Perceptron    (c) Support Vector Machines

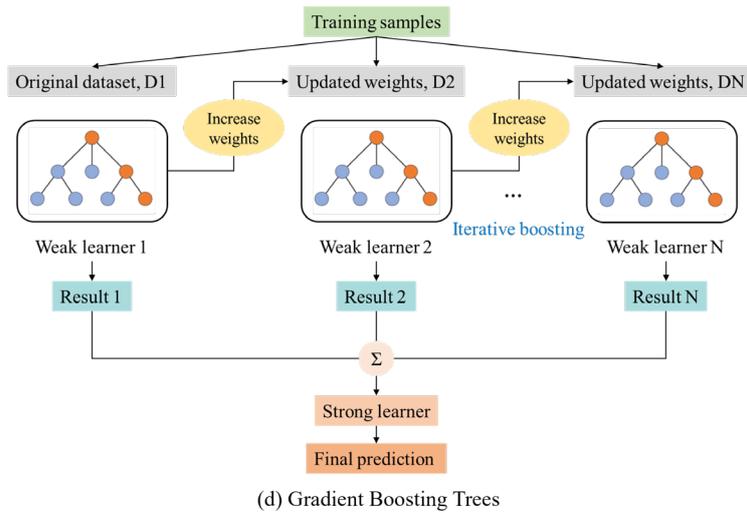

(d) Gradient Boosting Trees

Fig. 10. Overview of the structures of the multivariate regression models.

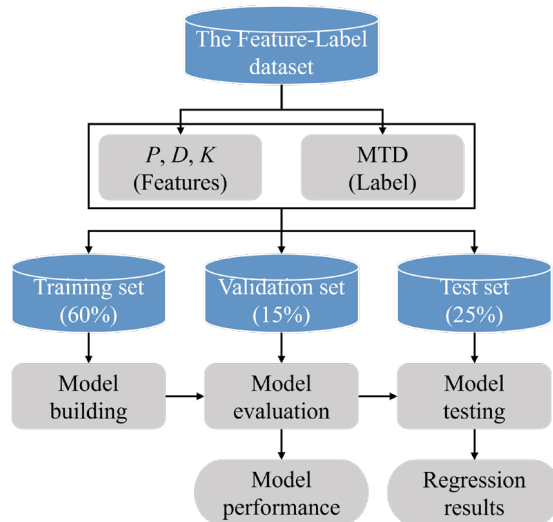

Fig. 11. Distribution of the dataset.

## 3. Results and discussion

### 3.1. Texture reconstruction results

Fig. 12 illustrates the results of depth map reconstruction for various pavements, all of which exhibit clear texture details. The camera coordinate system of the depth map is defined as shown in Fig. 13, where points farther away from the camera have relatively larger depth values stored in the pixels. Based on an understanding of this principle, it becomes easy to extract two-dimensional contours and three-dimensional features from the maps.

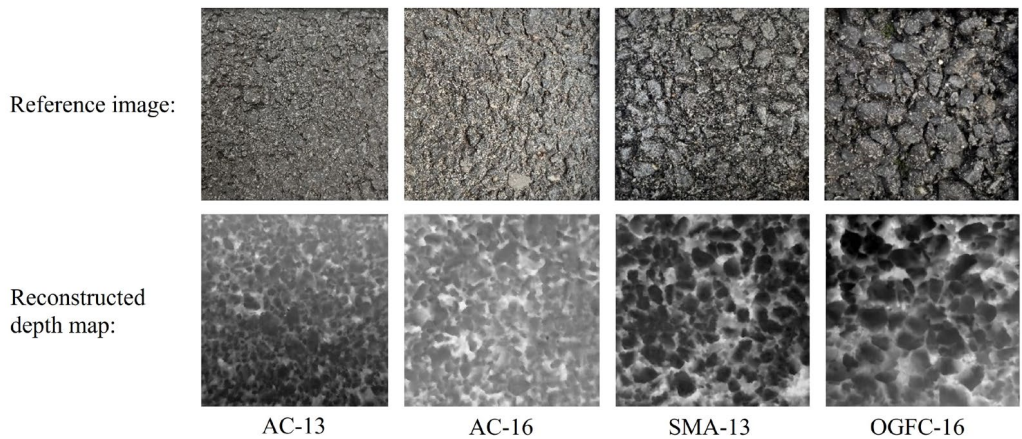

Fig. 12. Examples of reconstructed texture for different pavements.

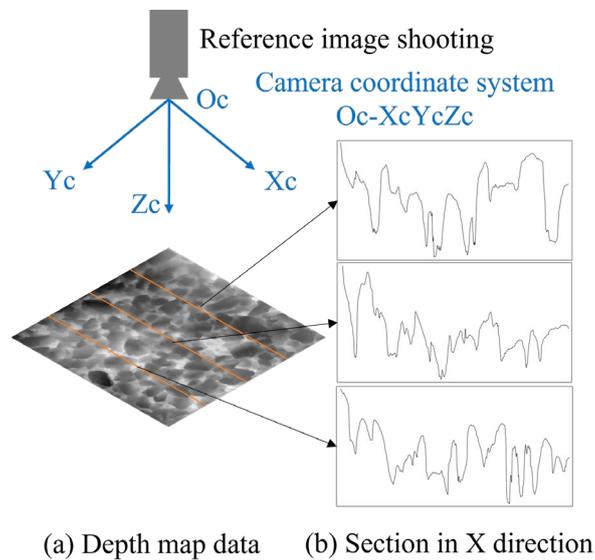

Fig. 13. Pavement profiles extracted from the depth map.

## 3.2. 3D data processing outcomes

### 3.2.1. Image filtering: A comparative analysis

As demonstrated in Fig. 14, the proposed filtering algorithm successfully preserves the original information in the map while effectively eliminating noise, without causing excessive smoothing. Some noise in the depth map manifests as discrete values or discontinuous regions, complicating the task of visually differentiating it from normal data. To address this, depth maps are converted into point clouds, as illustrated in Fig. 15. This conversion enables each point in a depth map to correspond to a 3D surface point, facilitating a more intuitive assessment of the filtering effect.

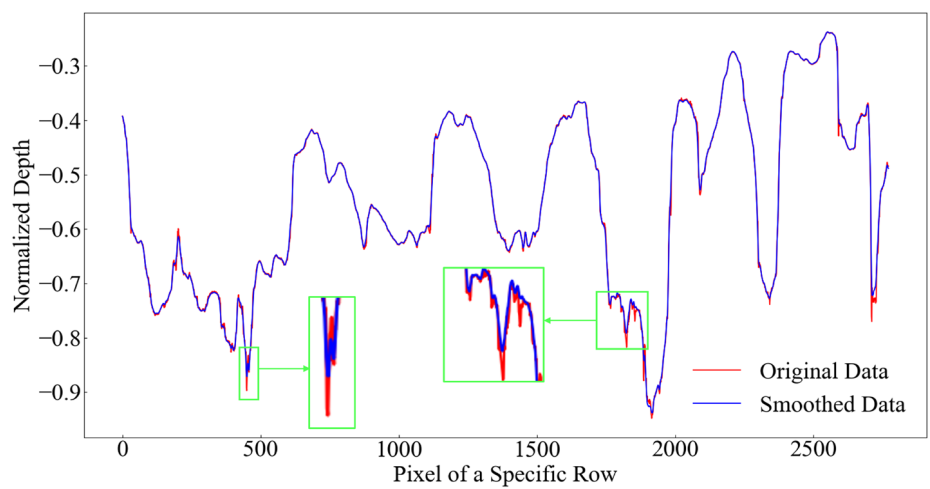

Fig. 14. Examples of texture data profiles before and after filtering.

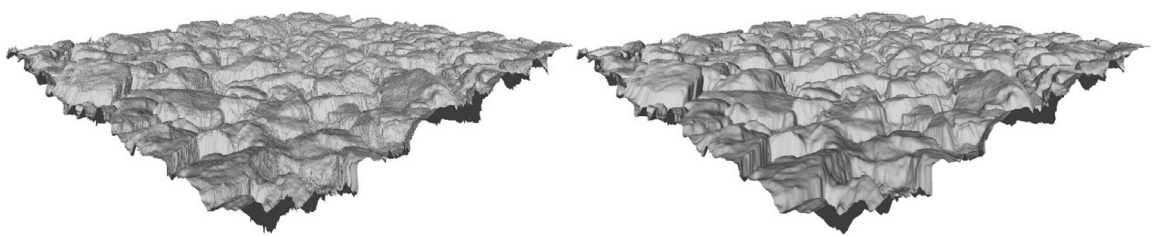

Fig. 15. Visualization of point clouds before and after filtering.

The Mean Square Error (*MSE*) is a commonly used metric for evaluating the effectiveness of filtering algorithms [84]. It quantifies the difference between the filtered and the original data, with a lower *MSE* indicating superior filtering performance. Although median filters are known to deliver improved filtering results as evidenced by the *MSE*s listed in Table 6, they can alter and distort the values of pavement data, as noted in [85]. This problem does not occur with the proposed method, which primarily relies on the bilateral filter as detailed in [37]. The *MSE* for the proposed algorithm is the lowest among the remaining methods evaluated, underscoring its exemplary performance. The equation for *MSE* is as follows:

$$MSE = \frac{1}{mn}\sum_{i=1}^{m}\sum_{j=1}^{n}|z(i,j) - z'(i,j)|^2 \qquad (14)$$

where $z(i,j)$ and $z'(i,j)$ denote the original data and the filtered data, respectively, and *m* and *n* denote the rows and columns of the depth map, respectively.

Table 6. *MSE*s of different filtering methods.

| Methods | Mean filter | Median filter | Bilateral filter | Proposed method |
|---|---|---|---|---|
| *MSE* | 2.4095e-05 | 1.4733e-05 | 8.1899e-05 | **2.4061e-05** |

*3.2.2. Tilt and offset correction results*

Fig. 16 presents the correction outcomes of a particular texture profile employing two distinct methods. The slope of the profile was determined using the least squares method to execute a linear fit across all data points. The results reveal that the method based on surface analysis is superior in mitigating errors. This superior performance is attributed to the fact that the correction technique, which relies on fitting a surface, succeeds to account for the local deformations present in the pavement. Fig. 17 displays the comparative outcomes prior to and following the application of data correction. In the corrected map, the variations in color are diminished, particularly noticeable in the lower-right corner where the colors more closely resemble those in the upper-left corner. This change suggests a reduction in the overall range of depth variations, further supporting the effectiveness of the introduced method.

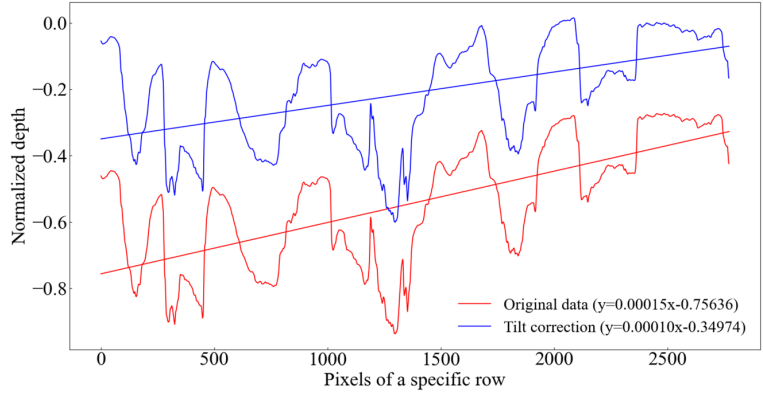

(a) The result of the calibration with the plane hypothesis.

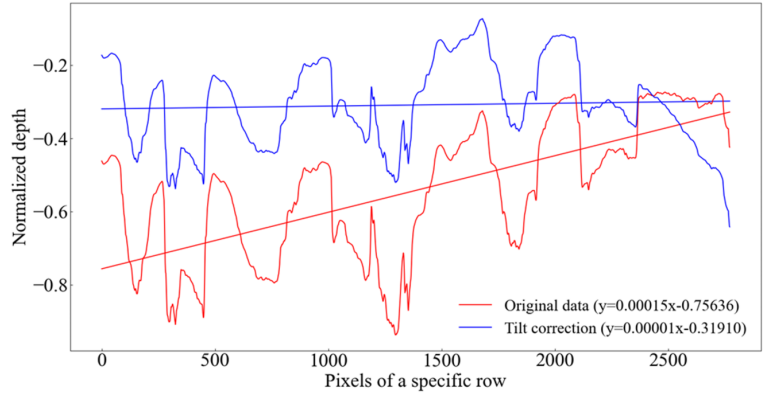

(b) The result of the calibration with the deformation hypothesis.

Fig. 16. Tilt and offset correction results using different methods.

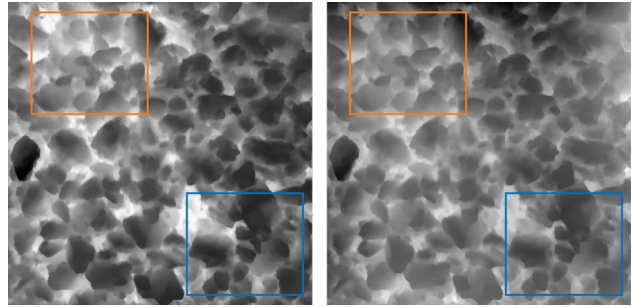

Fig. 17. Comparison of the depth map before and after correction.

### 3.3. Comprehensive analysis of the proposed features

The feature analysis results obtained from two distinct methods are separately detailed in Fig. 18 and Fig. 19. In the univariable linear regression analysis showcased in Fig. 18:

- Modeling with feature *P* results in a relatively high *SSE*, indicating that while there is a certain correlation between the concavity ratio and texture depth (Adjusted $R^2$=0.8268, p-value<0.0001), prediction errors arise due to the omission of other texture features or the possibility of a nonlinear relationship between *P* and MTD.

- Modeling with feature *K* yields a commendable regression result (Adjusted $R^2$=0.8312, p-value<0.0001), albeit with a higher *SSE* of 0.8464 than that observed in the multivariate regression model incorporating *P* and *D* (*SSE*=0.7160). The reasons for prediction errors are consistent with the previous point.
- Feature *D* demonstrates a moderate correlation with MTD (Adjusted $R^2$=0.4790), underscoring that aggregate size can provide reference information for MTD predictions. Within mixtures of the same type, it is observed that larger particle sizes correlate with increased MTD. For instance, the MTD of AC-16 typically surpasses that of AC-13. Nonetheless, the explanatory capacity of feature *D* regarding MTD is limited, as particle sizes might be comparable across different types of mixtures, yet the MTDs can exhibit significant variance. For example, the particle sizes of AC-13 and SMA-13 mixtures might not vary markedly, yet differences in their texture depth can be observed. Moreover, a p-value of 0.0076 also indicates that there isn't a direct linear relationship between the aggregate size and MTD.

For the analysis of multivariable linear regression shown in Fig. 18:

- Modeling that incorporates both *P* and *K* achieves an Adjusted $R^2$ of 0.8188, which does not perform as well as univariate regression models for either *P* or *K* alone. This is because the presence of multicollinearity between *P* and *K* (Adjusted $R^2$=0.9670), as illustrated in Fig. 19, diminishes their independent contributions to MTD prediction. Similarly, modeling with *P*, *K*, and *D* results in an Adjusted $R^2$ of 0.8217, which is also inferior to predictions made using *P* and *K* individually due to the same reason.
- Modeling with *P* and *D* yields the best result (Adjusted $R^2$=0.8413, *SSE*=0.716), indicating that incorporating additional information about aggregate size can enhance MTD prediction when concavity ratio is included. This is because *P* and *D* provide complementary texture information, thereby offering a more comprehensive reflection of MTD changes.
- Modeling with *K* and *D* provides less effective results (Adjusted $R^2$=0.8135, *SSE*=0.8416) compared to *P* and *D*, suggesting that the information provided by aggregate voids does not integrate as effectively with particle size information as the concavity ratio does in explaining variations in MTD. This indicates that *P* and *D* offer a greater volume of information for predicting MTD.

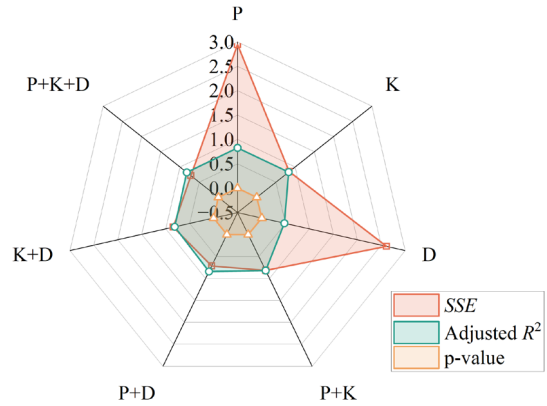

Fig. 18. The results of linear regression analysis.

As demonstrated in the correlation matrix shown in the Fig. 19, there is a high degree of linear correlation between *P* and *K*, with a *VIF* score of 30.3030. There is also a certain level of correlation between *P* and *D*, and between *K* and *D*, with their *VIF* scores being 4.3404 and 2.8736, respectively. In conclusion, while multivariate regression (*P+D*) shows fairly good predictive potential, establishing linear regression models that do not reach the ideal predictive performance (*SSE*≈0, Adjusted $R^2$>0.90, p-value<0.0001) indicates that the developed multivariate prediction models need to address multicollinearity among features [86, 87].

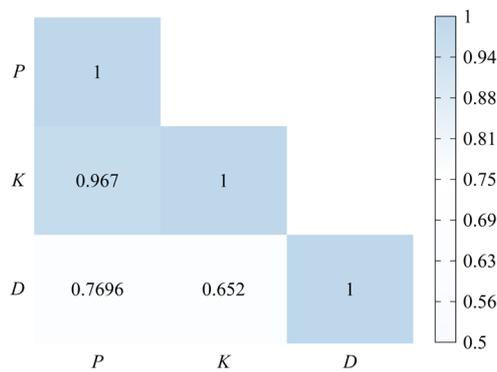

Fig. 19. The results of multicollinearity analysis.

*3.4. Multivariate regression model performance evaluation*

*3.4.1. Preliminary validation*

The RF, MLP, SVM, and GBT models were trained and validated separately. 96 of Feature-Label datasets are designated for model training, while 24 datasets were set aside for validation purposes. To ensure consistency in the feature distribution between the two datasets, a stratified random split was performed as shown in Fig. 20, aiming to maintain a similar sample proportion for each category in both datasets as in the original dataset [88-90]. The value range for MTD is between 0.55 to 1.15 mm, and 2.10 to 2.60 mm. It is worth noting that before the training and evaluation stages, the entire dataset was normalized using *Z*-score transformation to prevent input features with larger magnitudes from dominating those with smaller magnitudes [91, 92].

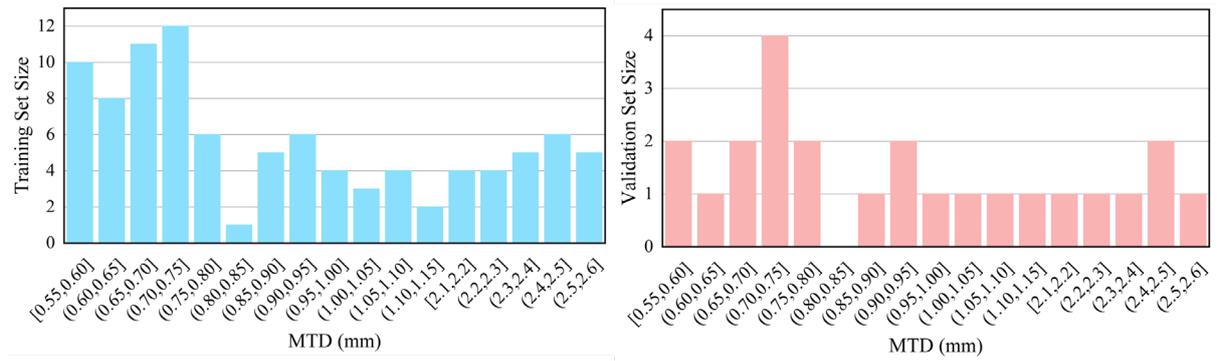

Fig. 20. Dataset sample distribution.

Four indicators are adopted to evaluate regressors' performances in the validation set: mean squared error (*MSE*), root mean square error (*RMSE*), mean absolute error (*MAE*), and $R^2$:

$$MSE = \frac{1}{n}\sum_{i=1}^{n}(y_i - \hat{y}_i)^2 \tag{15}$$

$$RMSE = \sqrt{MSE} \tag{16}$$

$$MSE = \frac{1}{n}\sum_{i=1}^{n}|y_i - \hat{y}_i| \tag{17}$$

$$R^2 = 1 - \frac{\sum_{i=1}^{n}(y_i - \hat{y}_i)^2}{\sum_{i=1}^{n}(y_i - \bar{y})^2} \qquad (18)$$

where $n$ is the number of samples, $y_i$ is the true label value, $\hat{y}_i$ is the predicted value, and $\bar{y}$ is the mean of the true label values. Table 7 shows the hyperparameters of the models. The implementation of machine learning algorithms requires configuring multiple tuning parameters [93-95], and the parameters shown are those that achieved the best performance on the validation set.

Table 7. The hyperparameters of different models.

| Model | Features | Parameters |
| --- | --- | --- |
| RF |  | RandomForestRegressor (n_estimators=60) |
| MLP | P+D | MLPRegressor (hidden_layer_sizes= (100, 50), activation='relu', |
|  | K+D | solver='adam', max_iter=1000) |
| SVM | P+K | SVR (kernel='poly') |
| GBT | P+K+D | GradientBoosting-Regressor (n_estimators=60, max_depth=5, learning_rate=0.1) |

Analysis of Fig. 21 indicates that the prediction performances of the RF, MLP, and SVM models are notably weak when $K + D$ are used as input features, evidenced by higher *MSE*, *RMSE*, *MAE* values, and lower $R^2$ scores. Although the GBT demonstrates a smaller prediction error with $K + D$ as input, its lower $R^2$ score compared to other inputs suggests a lack of critical feature information '*P*'. Consequently, the *K+D* input is deemed unsuitable for final model consideration. When *P+K+D*, *P+D*, or *P+K* are used as inputs, the prediction performance across the models is relatively similar, which warrants the use of cross-validation to further assess the generalization ability and stability of the models [96-98]. It's worth noting that the GBT yielded identical prediction results when the inputs were *P+D* and *P+K+D*. This occurred because the GBT selects the optimal features for constructing tree nodes in each training iteration [99, 100]. When additional features (in this case, *K*) do not provide extra predictive power on a given dataset, the model tends to choose the same features (in this case, *P+D*) for prediction [101, 102]. As a result, the input of *P+K+D* will not be used in further evaluations of the GBT to minimize the number of inputs as much as possible. Furthermore, the RF exhibits significantly higher prediction errors and lower fit when using *P+D* as input, leading to the exclusion of *P+D* input from the RF's cross-validation process.

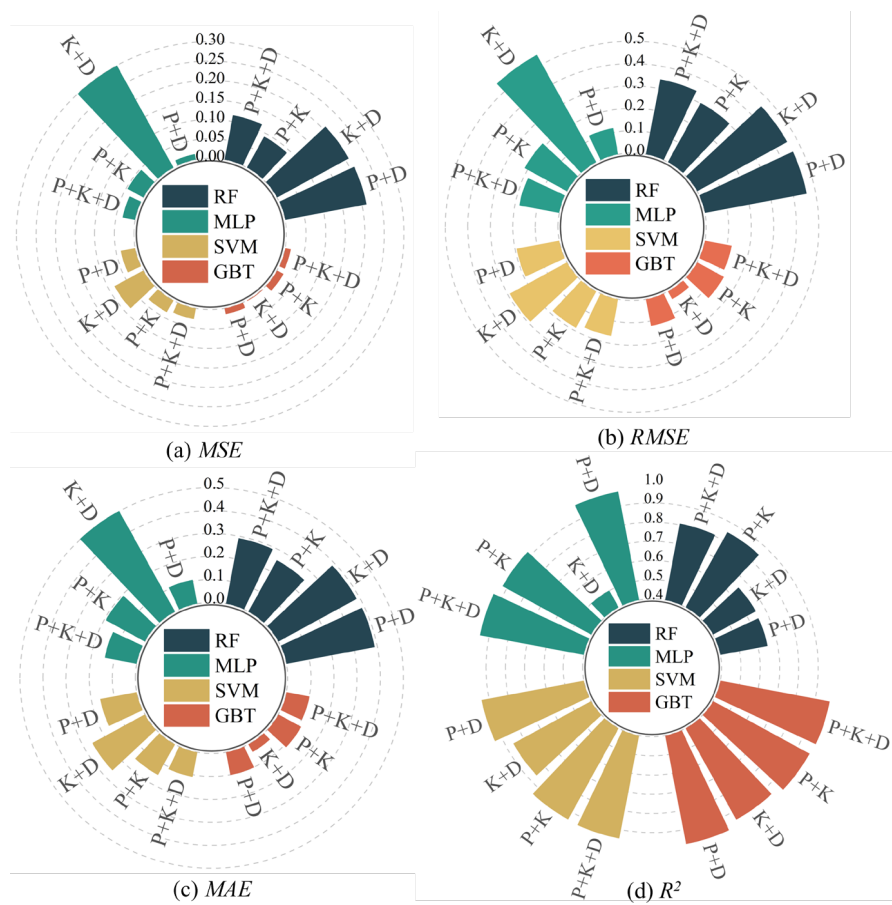

(a) *MSE*     (b) *RMSE*     (c) *MAE*     (d) $R^2$

Fig. 21. Regression performances of the multivariate prediction models.

### 3.4.2. Cross validation

The generalization ability and stability of the models were assessed using K-fold cross-validation [103-105]. Inspired by [106], the 120 datasets were divided into 5 folds, with one fold serving as the validation set in each iteration, while the remaining folds were used for training. This procedure was repeated 5 times. The $R^2$ results obtained from the process, using Equation (18), are graphically represented by the box plot in Fig. 22. An $R^2$ value close to 1 indicates a high goodness of fit for the model [107, 108]. It can be observed that the median $R^2$ values of the RF models across the 5 experiments are below 0.85, significantly inferior in terms of generalization ability compared to others. The whiskers of both RF and MLP models are relatively long, indicating the presence of extreme $R^2$ values in the fitting results that significantly deviate from the medians. This suggests they cannot maintain consistent fitting performance across multiple predictions [109, 110]. Notably, there are small outliers far from the main body of the data in MLP (*P+K*) and MLP (*P+K+D*), indicating potential poorer fitting results when these models are used to predict MTD. Additionally, the wider box plots of the RF and MLP models indicate greater fluctuations of $R^2$ around the medians, suggesting instability in their predictions [111, 112]. The $R^2$ medians of the SVM and GBT models are not significantly different, both being around 0.95, which demonstrates good generalization ability. However, the GBT model exhibits better predictive stability in terms of the narrower width of the boxes and shorter whiskers. In summary, the GBT model outperforms others in terms of both generalization ability and stability.

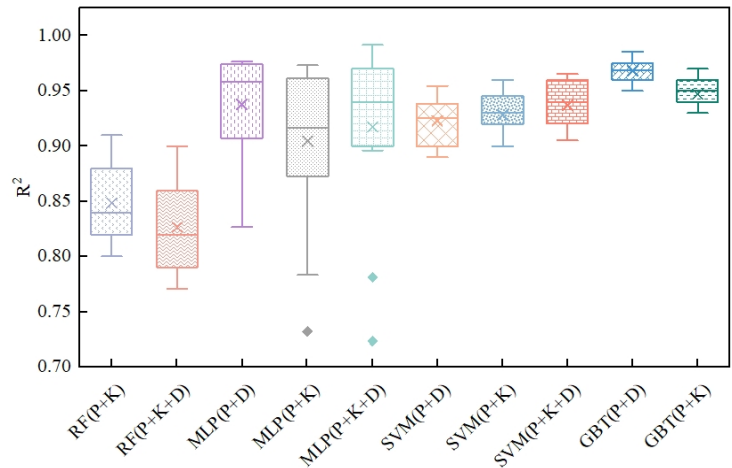

Fig. 22. The $R^2$ results of cross-validation.

To demonstrate the GBT model's superiority and select the most suitable input features, an accuracy assessment was conducted comparing SVM and GBT models. This assessment utilized the linear fitting equations of MTD prediction values and label values. The closer the slope is to 1 and the intercept to 0, the closer the predicted values are to the baseline values, indicating a model's better predictive accuracy [113, 114]. The results, as illustrated in Fig. 23, show that the SVM models exhibit greater variability in the five calculations of slope and intercept, leading to less stable predictive accuracy compared to the GBT models. The average values of the slope and intercept for these models are presented in Fig. 24. Despite the slight differences in average slopes across the models, approximately around 0.90, the GBT models demonstrate smaller average intercepts, with the most accurate predictions observed when using *P* and *D* as input features. Therefore, considering the model's generalizability, stability, and accuracy, the GBT model, with *P* and *D* as its input, is selected as the prediction model. Fig. 25 visualizes the cross-validation results of the GBT model when using *P+D* as input. The average values of the linear regression equations' $R^2$, slope, and intercept are 0.9858, 0.9126, and 0.07268 respectively.

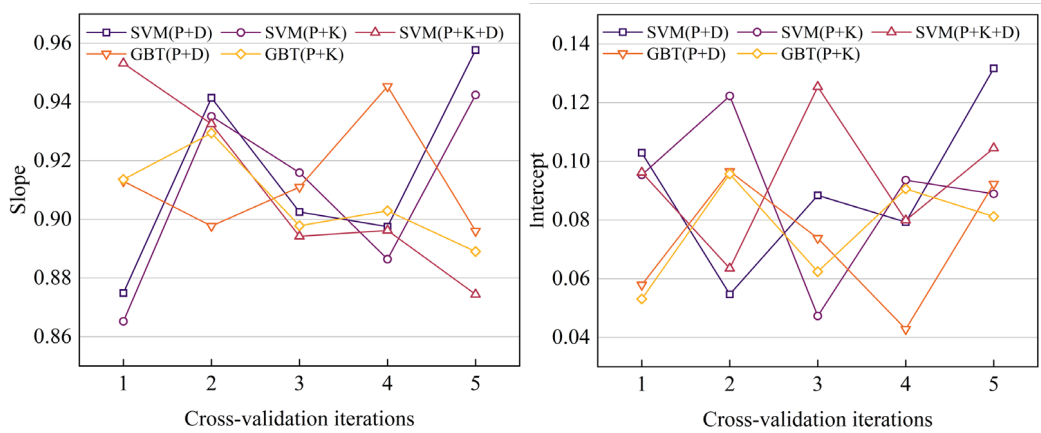

Fig. 23. The results of linear equation fitting.

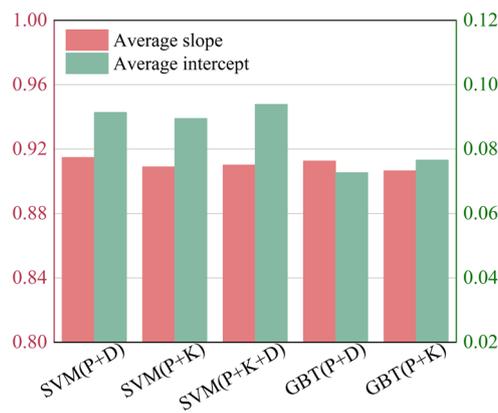

Fig. 24. Model accuracy evaluation results.

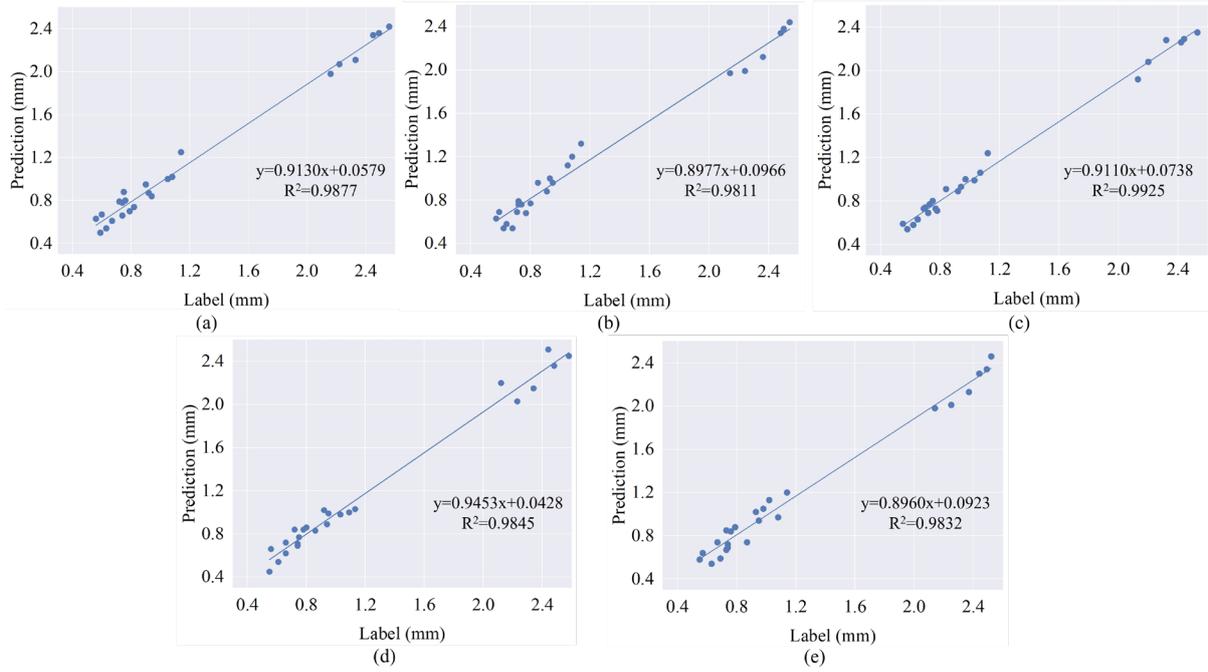

Fig. 25. Regression results of the established prediction model.

## 3.5. Comparative field testing of different techniques

To assess the effectiveness of the proposed method and to compare it with various MTD measurement techniques, field experiments were carried out using 40 sets of Feature-Label datasets as the test data. The evaluation included the proposed method alongside other MTD assessment techniques: (1) Conversion of the reference image to an 8-bit grayscale image using the standard mean gray-level method, where the gray level difference indicative of surface height variation was utilized for MTD prediction, as discussed in reference [8]; (2) Application of the 3D reconstruction method and computational model from study [13], generating point clouds based on the Structure from Motion (SfM) principle for MTD calculation; (3) Implementation of the method described in Section 2.1 for pavement texture reconstruction, with depth information calibrated and extracted using the approach from study [14] to assess MTD. These methods are referred to as the 3D Image-Based Texture Analysis Method (3D-ITAM), the Structure from Motion (SfM) technique, and the Deep Learning (DL) method requiring calibration, respectively. They were described in more detail in Section 1.

From Fig. 26, it is clear that the fitting performance of the proposed method, with an $R^2$ value of 0.9827, ranks second, performing better than both the 3D-ITAM method and the DL method, and is not significantly different from the SfM method, which has the highest $R^2$ value of 0.9890. Examining the distribution of data points, the proposed method shows better fitting results for data points with smaller MTDs compared to those with larger MTDs. In the 3D-ITAM method, many data points are scattered on both sides of the fitting line, especially for measurements at smaller MTDs, resulting in the worst fitting effect. The SfM and DL methods perform better on data with larger MTDs compared to the proposed method, but not as well on data with smaller MTDs. Looking at the slope of the fitting equation, the proposed method has the highest value of 0.9423, indicating the greatest effect of the predictive variable on the response variable, i.e., the best linear relationship between predicted values and baseline values. Therefore, the fitting quality of the proposed method is relatively high, offering good measurement results for pavements with smaller MTDs. However, its linear correlation weakens for pavements with larger MTDs.

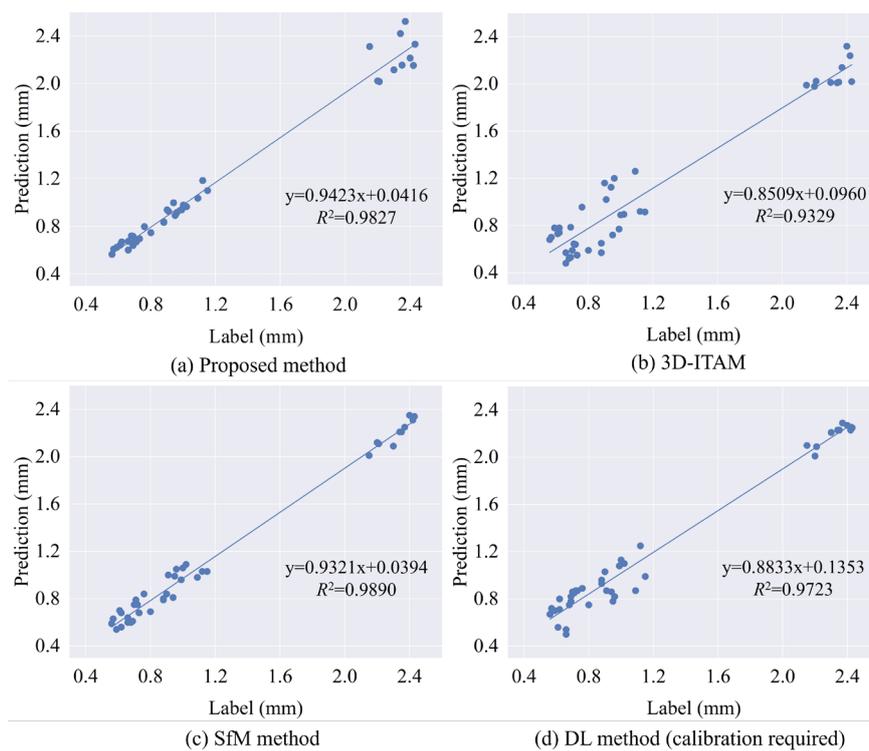

Fig. 26. The linear fitting results between the predicted values and the baseline values.

Fig. 27 shows the absolute and relative errors of various measurement methods, Analysis reveals that:

- The method proposed demonstrates accurate measurement results on AC-13, AC-16, and SMA-13, with the median absolute error not exceeding 0.05 mm. However, the testing results on OGFC-16 differ from baseline values obtained using the sand patch method, with a median absolute error of approximately 0.18 mm. This discrepancy is because image-based methods cannot capture some deep voids with small openings in porous hydrophobic pavements like OGFC, whereas the standard sand can fill these voids, resulting in lower MTD predictions. The box for relative error on the AC-13 is relatively long, indicating a certain volatility in the data. This is because the MTDs of AC-13 is relatively small, but there exists variation in MTDs among samples (MTDs for AC-13 range from 0.55 to 0.79 mm). Consequently, the relatively small absolute error can result in substantially different relative errors.
- The 3D-ITAM is relatively inaccurate, with both its absolute and relative errors being higher across various pavements compared to the proposed method. The presence of long whiskers and wide boxes on all pavements indicates a wide range of error fluctuations, suggesting the unpredictability of the 3D-ITAM. This is because it only utilizes information from a single image for texture feature extraction, resulting in poorer accuracy compared to the proposed method, where features are acquired from 3D data generating from multi-view images. Moreover, the 3D-ITAM converts digital images into 8-bit precision grayscale images, losing many fine texture details compared to the proposed method's 32-bit precision 3D data. Furthermore, its prediction for MTD relies on fitted empirical models, which lack interpretability and therefore its usage is not recommended.
- The SfM method shows higher absolute errors on AC-13, AC-16, and SMA-13 compared to the proposed method. On AC-16 and SMA-13, the range of relative error fluctuations is larger than that of the proposed method. This is because the reconstruction model used by the SfM method has not been trained with extensive pavement data, leading to the generated texture data being inferior in accuracy to depth data obtained from the DL model used in the proposed method, which results in a less accurate prediction of MTD values. However, on OGFC surfaces, both the absolute and relative errors of the SfM method are lower than those of the proposed method, indicating that the point clouds generated by the SfM method are better at characterizing porous surfaces. Nonetheless, the absolute error of the SfM method on OGFC exhibits fluctuations, making it difficult to guarantee consistently high measurement accuracy.

- The DL method that requires calibration shows higher relative and absolute errors on AC-13, AC-16, and SMA-13 than both the proposed method and the SfM method. However, its measurement errors are still smaller than those of the 3D-ITAM, indicating the reliability of obtaining 3D data based on the DL model. The magnitude of errors in this DL method mainly depends on the accuracy of the calibration points selected. Particularly for surfaces like AC-13, which have smaller MTDs, unless the calibration points are very precise, converting relative depth maps into absolute depth maps can easily lead to significant errors, ultimately resulting in inaccurate MTD predictions. In contrast, the proposed method predicts MTD by extracting texture features from relative depth maps, thereby avoiding the need for calibration and enhancing the reliability of image-based measurement methods. It is noted that this DL method predicts the texture depth of OGFC more accurately than the proposed method, suggesting that image-based methods have the potential to achieve more accurate MTD predictions on OGFC. The feature extraction method designed in this paper has not fully considered the texture information of such porous surfaces, indicating the need for future research in this area.

In summary, the presented approach showcases its accuracy and consistency in evaluations across AC-13, AC-16, and SMA-16. While its accuracy on OGFC-16 requires enhancement, its stability remains unmatched when compared to alternative techniques. The median relative error of this method remains below 8%, with the maximum relative errors not surpassing 10%, underscoring its overall dependability. In terms of automation, the proposed method is fully implemented using a custom-developed Python algorithm, enabling end-to-end image input and MTD prediction data output. Unlike SfM methods, this approach does not entail following multiple software operation steps to build a 3D point cloud model. Additionally, it also eliminates the necessity to switch between various image processing and data extraction software, as compared to the 3D-ITAM. Moreover, it eradicates the need for manual calibration of depth maps, thereby showing potential for complete automation.

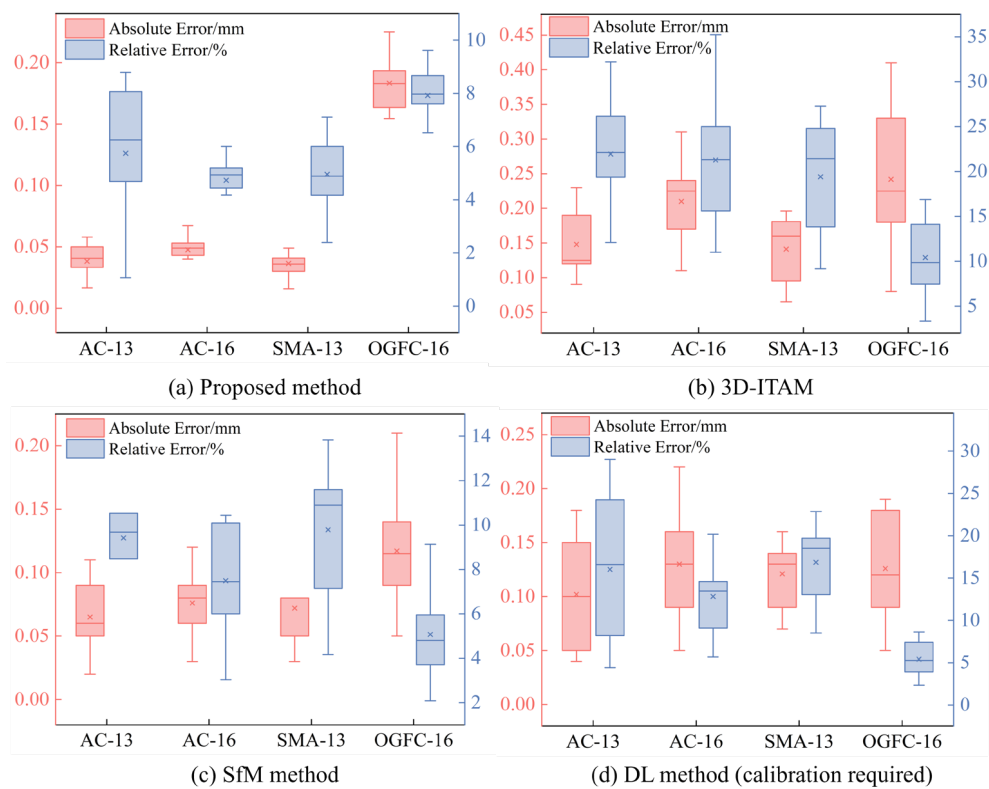

Fig. 27. Measurement errors of different methods.

## 4. Conclusive insights

The detection of pavements using deep learning and machine learning has become a focal point of research, particularly with the growing popularity of utilizing 3D data. However, challenges such as the high cost of sensor equipment and complex operational procedures persist. This article introduces an automated method for determining pavement texture depth, requiring only a digital camera and a computer. The primary contributions of this study are as follows:

- High-precision depth map data is produced using an image-based multi-view 3D reconstruction deep learning model named Patchmatchnet, which utilizes a cost-effective camera instead of expensive laser devices for 3D data generation.
- An adaptive local bilateral filtering algorithm is proposed, preserving original 3D data features more effectively than traditional methods. Additionally, a surface-fitting-based tilt correction method is introduced, which proves more practical than the plane-fitting-based method when considering pavement deformation.

- Three features—namely, the relative concavity ratio ($P$), the maximum particle size ratio ($D$), and the aggregate voids ($K$)—that influence the texture depth from various aspects are extracted from depth map data. An analysis of the relationship between features and MTD demonstrates the interpretability of how these features influence texture depth.

- Multivariate regressors, utilizing the proposed features as input, have been established for MTD prediction. The Gradient Boosting Trees model achieves an optimal balance between prediction accuracy and stability, with an $R^2$ value of 0.9858 in cross-validation.

- Compared to other techniques, the proposed method demonstrates overall optimal performance across various types of pavements, with relative errors from field tests being less than 10%.

The proposed method enables an automated and end-to-end assessment of pavement skid resistance, saving manpower and time for pavement management systems. However, some issues need to be addressed through future technical iterations:

- The image-based 3D texture data acquisition method is susceptible to environmental lighting and pavement cleanliness, necessitating improvements in the 3D reconstruction model's structure to recover the true texture appearance even under environmental impacts.

- The accuracy of MTD prediction for OGFC pavements needs improvement. Future work should develop features more suitable for characterizing porous pavements and other special material surfaces.

- The multivariate regression model's training and validation are based on data from four types of mixtures. Predicting the texture depth of other mixtures may not be accurate. Future efforts should focus on expanding the Feature-Label dataset to adapt the prediction model to more pavement types.

**Disclosure statement**

The authors declare that they have no known competing financial interests or personal relationships that could have appeared to influence the work reported in this paper.

**Data availability**

The data that support the findings of this study are available on request from the corresponding author. The data are not publicly available due to privacy restrictions.

**Funding**

This work was supported by the National Natural Science Foundation of China (grant numbers 52278468 and U22A20235); the Hunan Transportation Science and Technology Foundation (CN) (grant number 202104); and Science and Technology Research and Development Program of China Railway Group Limited (Grant No: 2022-ZD-13).

**Data availability**

The data that support the findings of this study are available on request from the corresponding author. The data are not publicly available due to privacy restrictions.

**Funding**

This work was supported by the National Natural Science Foundation of China (grant numbers 52278468 and U22A20235); the Hunan Transportation Science and Technology Foundation (CN) (grant number 202104); and Science and Technology Research and Development Program of China Railway Group Limited (Grant No: 2022-ZD-13).

**References**

[1] Q. Liu, A. Shalaby, Relating concrete pavement noise and friction to three-dimensional texture parameters, International Journal of Pavement Engineering 18(5) (2017) 450-458.

[2] H. Dan, L. Gao, H. Wang, J. Tang, Discrete-element modeling of mean texture depth and wearing behavior of asphalt mixture, Journal of Materials in Civil Engineering 34(4) (2022) 04022027.

[3] C. Liu, Y. Du, S. Wong, G. Chang, S. Jiang, Eco-based pavement lifecycle maintenance scheduling optimization for equilibrated networks, Transportation Research Part D: Transport and Environment 86 (2020) 102471.

[4] T. Hu, W. Zhu, Y. Yan, Artificial intelligence aspect of transportation analysis using large scale systems, Proceedings of the 2023 6th Artificial Intelligence and Cloud Computing Conference, 2023, pp. 54-59.

[5] S. Torbruegge, B. Wies, Characterization of pavement texture by means of height difference correlation and relation to wet skid resistance, Journal of traffic and transportation engineering (English edition) 2(2) (2015) 59-67.

[6] K. Gao, H. Zhang, G. Wu, A multispectral vision-based machine learning framework for non-contact vehicle weigh-in-motion, Measurement 226 (2024) 114162.

[7] A. El Gendy, A. Shalaby, M. Saleh, G.W. Flintsch, Stereo-vision applications to reconstruct the 3D texture of pavement surface, International Journal of Pavement Engineering 12(03) (2011) 263-273.

[8] D. Yu, Y. Cao, Q. Zhao, Detection and Analysis of Asphalt Pavement Texture Depth Based on Digital Image Analytics Technology, International Journal of Pavement Research and Technology (2023) 1-10.

[9] Z. Weng, H. Xiang, Y. Lin, C. Liu, D. Wu, Y. Du, Pavement texture depth estimation using image-based multiscale features, Automation in Construction 141 (2022) 104404.

[10] R.B. Kogbara, E.A. Masad, D. Woodward, P. Millar, Relating surface texture parameters from close range photogrammetry to Grip-Tester pavement friction measurements, Construction and Building Materials 166 (2018) 227-240.

[11] Z. Lu, Y. Chen, Self-supervised monocular depth estimation on water scenes via specular reflection prior, Digital Signal Processing 149 (2024) 104496.

[12] D. Ma, Y. Yang, Q. Tian, B. Dang, Z. Qi, A. Xiang, Comparative analysis of x-ray image classification of pneumonia based on deep learning algorithm, Theoretical and Natural Science 56 (2024) 52-59.

[13] H.-C. Dan, G.-W. Bai, Z.-H. Zhu, X. Liu, W. Cao, An improved computation method for asphalt pavement texture depth based on multiocular vision 3D reconstruction technology, Construction and Building Materials 321 (2022) 126427.

[14] H.-C. Dan, B. Lu, M. Li, Evaluation of asphalt pavement texture using multiview stereo reconstruction based on deep learning, Construction and Building Materials 412 (2024) 134837.

[15] Z. Lu, Y. Chen, Pyramid frequency network with spatial attention residual refinement module for monocular depth estimation, Journal of Electronic Imaging 31(2) (2022) 023005-023005.

[16] X. Cui, X. Zhou, J. Lou, J. Zhang, M. Ran, Measurement method of asphalt pavement mean texture depth based on multi-line laser and binocular vision, International Journal of Pavement Engineering 18(5) (2017) 459-471.

[17] Y. Wang, B. Yu, X. Zhang, J. Liang, Automatic extraction and evaluation of pavement three-dimensional surface texture using laser scanning technology, Automation in Construction 141 (2022) 104410.

[18] X. Wang, C. Wang, B. Liu, X. Zhou, L. Zhang, J. Zheng, X. Bai, Multi-view stereo in the deep learning era: A comprehensive review, Displays 70 (2021) 102102.

[19] X. Zhu, Y. Zhu, H. Wang, H. Wen, Y. Yan, P. Liu, Skeleton sequence and RGB frame based multi-modality feature fusion network for action recognition, ACM Transactions on Multimedia Computing, Communications, and Applications (TOMM) 18(3) (2022) 1-24.

[20] P. Dong, Z. Kong, X. Meng, P. Yu, Y. Gong, G. Yuan, H. Tang, Y. Wang, HotBEV: Hardware-oriented transformer-based multi-view 3D detector for BEV perception, Advances in Neural Information Processing Systems 36 (2024).

[21] D. Zhang, P. Zhi, B. Yong, J.-Q. Wang, Y. Hou, L. Guo, Q. Zhou, R. Zhou, EHSS: An Efficient Hybrid-supervised Symmetric Stereo Matching Network, 2023 IEEE 26th International Conference on Intelligent Transportation Systems (ITSC), IEEE, 2023, pp. 1044-1051.

[22] X. Bu, Y. Wu, Z. Gao, Y. Jia, Deep convolutional network with locality and sparsity constraints for texture classification, Pattern Recognition 91 (2019) 34-46.

[23] J. Zhuang, D. Wang, Geometrically matched multi-source microscopic image synthesis using bidirectional adversarial networks, Proceedings of 2021 International Conference on Medical Imaging and Computer-Aided Diagnosis (MICAD 2021) Medical Imaging and Computer-Aided Diagnosis, Springer, 2022, pp. 79-88.

[24] G. Liu, J. Qian, F. Wen, X. Zhu, R. Ying, P. Liu, Action recognition based on 3d skeleton and rgb frame fusion, 2019 IEEE/RSJ International Conference on Intelligent Robots and Systems (IROS), IEEE, 2019, pp. 258-264.

[25] F. Shen, H. Ye, J. Zhang, C. Wang, X. Han, W. Yang, Advancing pose-guided image synthesis with progressive conditional diffusion models, arXiv preprint arXiv:2310.06313 (2023).

[26] Z. Wang, H. Yu, X. Zhu, Z. Li, C. Chen, L. Song, Learning 3D Human Pose and Shape Estimation Using Uncertainty-Aware Body Part Segmentation, ICASSP 2023-2023 IEEE International Conference on Acoustics, Speech and Signal Processing (ICASSP), IEEE, 2023, pp. 1-5.

[27] M. Li, S. Liu, H. Zhou, G. Zhu, N. Cheng, T. Deng, H. Wang, Sgs-slam: Semantic gaussian splatting for neural dense slam, European Conference on Computer Vision, Springer, 2025, pp. 163-179.

[28] T. Jiang, L. Liu, J. Jiang, T. Zheng, Y. Jin, K. Xu, Trajectory tracking using frenet coordinates with deep deterministic policy gradient, arXiv preprint arXiv:2411.13885 (2024).

[29] Z. Lu, Y. Chen, Joint self-supervised depth and optical flow estimation towards dynamic objects, Neural Processing Letters 55(8) (2023) 10235-10249.

[30] T. Deng, Y. Wang, H. Xie, H. Wang, J. Wang, D. Wang, W. Chen, Neslam: Neural implicit mapping and self-supervised feature tracking with depth completion and denoising, arXiv preprint arXiv:2403.20034 (2024).

[31] F. Shen, J. Tang, Imagpose: A unified conditional framework for pose-guided person generation, The Thirty-eighth Annual Conference on Neural Information Processing Systems, 2024.

[32] Z. Wang, B. Li, C. Wang, S. Scherer, AirShot: Efficient Few-Shot Detection for Autonomous Exploration, arXiv preprint arXiv:2404.05069 (2024).

[33] J. Tang, W. Zhang, H. Liu, M. Yang, B. Jiang, G. Hu, X. Bai, Few could be better than all: Feature sampling and grouping for scene text detection, Proceedings of the IEEE/CVF Conference on Computer Vision and Pattern Recognition, 2022, pp. 4563-4572.

[34] J. Tang, W. Qian, L. Song, X. Dong, L. Li, X. Bai, Optimal boxes: boosting end-to-end scene text recognition by adjusting annotated bounding boxes via reinforcement learning, European Conference on Computer Vision, Springer, 2022, pp. 233-248.

[35] Y. Chen, S. Yan, Z. Zhu, Z. Li, Y. Xiao, Xmecap: Meme caption generation with sub-image adaptability, Proceedings of the 32nd ACM International Conference on Multimedia, 2024, pp. 3352-3361.

[36] Z. Wei, Y. Huang, Y. Chen, C. Zheng, J. Gao, A-ESRGAN: Training real-world blind super-resolution with attention U-Net Discriminators, Pacific Rim International Conference on Artificial Intelligence, Springer, 2023, pp. 16-27.

[37] M.M. Ibrahim, Q. Liu, R. Khan, J. Yang, E. Adeli, Y. Yang, Depth map artefacts reduction: A review, IET Image Processing 14(12) (2020) 2630-2644.

[38] M.A. Fischler, R.C. Bolles, Random sample consensus: a paradigm for model fitting with applications to image analysis and automated cartography, Communications of the ACM 24(6) (1981) 381-395.

[39] J. Hampp, R. Bormann, Quadtree-based polynomial polygon fitting, 2013 IEEE/RSJ International Conference on Intelligent Robots and Systems, IEEE, 2013, pp. 4207-4213.

[40] J. Xiang, J. Chen, Imitation Learning-Based Convex Approximations of Probabilistic Reachable Sets, AIAA AVIATION FORUM AND ASCEND 2024, 2024, p. 4356.

[41] J. Yin, Z. Zheng, Y. Pan, Y. Gu, Y. Chen, Semi-supervised semantic segmentation with multi-reliability and multi-level feature augmentation, Expert Systems with Applications 233 (2023) 120973.

[42] Y. Chen, X. Meng, Y. Wang, S. Zeng, X. Liu, Z. Xie, LUCIDA: Low-Dose Universal-Tissue CT Image Domain Adaptation for Medical Segmentation, International Conference on Medical Image Computing and Computer-Assisted Intervention, Springer, 2024, pp. 393-402.

[43] J. Liu, Z. Kong, P. Zhao, W. Zeng, H. Tang, X. Shen, C. Yang, W. Zhang, G. Yuan, W. Niu, TSLA: A Task-Specific Learning Adaptation for Semantic Segmentation on Autonomous Vehicles Platform, IEEE Transactions on Computer-Aided Design of Integrated Circuits and Systems  (2024).

[44] A. Rezaei, E. Masad, A. Chowdhury, Development of a model for asphalt pavement skid resistance based on aggregate characteristics and gradation, Journal of transportation engineering 137(12) (2011) 863-873.

[45] J. Zhuang, M. Gao, M.A. Hasan, Lighter U-net for segmenting white matter hyperintensities in MR images, Proceedings of the 16th EAI International Conference on Mobile and Ubiquitous Systems: Computing, Networking and Services, 2019, pp. 535-539.

[46] Y. Chen, Y. Gao, L. Zhu, W. Shao, Y. Lu, H. Han, Z. Xie, PCNet: Prior Category Network for CT Universal Segmentation Model, IEEE Transactions on Medical Imaging (2024).

[47] J. Yin, S. Yan, T. Chen, Y. Chen, Y. Yao, Class Probability Space Regularization for semi-supervised semantic segmentation, Computer Vision and Image Understanding 249 (2024) 104146.

[48] T.R. Singh, S. Roy, O.I. Singh, T. Sinam, K.M. Singh, A new local adaptive thresholding technique in binarization, arXiv preprint arXiv:1201.5227 (2012).

[49] C. Oliviero Rossi, B. Teltayev, R. Angelico, Adhesion promoters in bituminous road materials: A review, Applied Sciences 7(5) (2017) 524.

[50] C.A. Schneider, W.S. Rasband, K.W. Eliceiri, NIH Image to ImageJ: 25 years of image analysis, Nature methods 9(7) (2012) 671-675.

[51] J. Liu, S. Chen, Q. Liu, Y. Wang, B. Yu, Influence of steel slag incorporation on internal skeletal contact characteristics within asphalt mixture, Construction and Building Materials 352 (2022) 129073.

[52] R.B. Kogbara, E.A. Masad, E. Kassem, A.T. Scarpas, K. Anupam, A state-of-the-art review of parameters influencing measurement and modeling of skid resistance of asphalt pavements, Construction and Building Materials 114 (2016) 602-617.

[53] L. Wang, L. Carvalho, Deviance matrix factorization, Electronic Journal of Statistics 17(2) (2023) 3762-3810.

[54] Y. Xiao, E. Sun, Y. Jin, Q. Wang, W. Wang, Proteingpt: Multimodal llm for protein property prediction and structure understanding, arXiv preprint arXiv:2408.11363 (2024).

[55] D. Luo, Optimizing Load Scheduling in Power Grids Using Reinforcement Learning and Markov Decision Processes, arXiv preprint arXiv:2410.17696 (2024).

[56] T. Tian, J. Deng, B. Zheng, X. Wan, J. Lin, AI-Driven Transformation: Revolutionizing Production Management with Machine Learning and Data Visualization, Journal of Computational Methods in Engineering Applications (2024) 1-18.

[57] P. Zhao, L. Lai, L. Shen, Q. Li, J. Wu, Z. Liu, A Huber Loss Minimization Approach to Mean Estimation under User-level Differential Privacy, arXiv preprint arXiv:2405.13453 (2024).

[58] X. Xu, Z. Wang, Y. Zhang, Y. Liu, Z. Wang, Z. Xu, M. Zhao, H. Luo, Style Transfer: From Stitching to Neural Networks, 2024 5th International Conference on Big Data & Artificial Intelligence & Software Engineering (ICBASE), IEEE, 2024, pp. 526-530.

[59] L. Wang, I. Lauriola, A. Moschitti, Accurate training of web-based question answering systems with feedback from ranked users, Proceedings of the 61st Annual Meeting of the Association for Computational Linguistics (Volume 5: Industry Track), 2023, pp. 660-667.

[60] D. Luo, Enhancing Smart Grid Efficiency through Multi-Agent Systems: A Machine Learning Approach for Optimal Decision Making, Preprints preprints 202411 (2024) v1.

[61] C. Peng, D. Zhang, U. Mitra, Graph Identification and Upper Confidence Evaluation for Causal Bandits with Linear Models, ICASSP 2024-2024 IEEE International Conference on Acoustics, Speech and Signal Processing (ICASSP), IEEE, 2024, pp. 7165-7169.

[62] K. Xu, L. Chen, J.-M. Patenaude, S. Wang, Kernel representation learning with dynamic regime discovery for time series forecasting, Pacific-Asia Conference on Knowledge Discovery and Data Mining, Springer, 2024, pp. 251-263.

[63] P. Yu, X. Xu, J. Wang, Applications of Large Language Models in Multimodal Learning, Journal of Computer Technology and Applied Mathematics 1(4) (2024) 108-116.

[64] L. Breiman, Random forests, Machine learning 45 (2001) 5-32.

[65] F. Rosenblatt, Principles of neurodynamics: Perceptrons and the theory of brain mechanisms, Spartan books Washington, DC1962.

[66] D. Zhang, S. Sen, The Stochastic Conjugate Subgradient Algorithm For Kernel Support Vector Machines, arXiv preprint arXiv:2407.21091 (2024).

[67] A. Natekin, A. Knoll, Gradient boosting machines, a tutorial, Frontiers in neurorobotics 7 (2013) 21.

[68] Q. Yang, P. Li, X. Xu, Z. Ding, W. Zhou, Y. Nian, A comparative study on enhancing prediction in social network advertisement through data augmentation, 2024 4th

International Conference on Machine Learning and Intelligent Systems Engineering (MLISE), IEEE, 2024, pp. 214-218.

[69] G. Wang, Z. Wang, P. Sun, A. Boukerche, SK-SVR-CNN: A Hybrid Approach for Traffic Flow Prediction with Signature PDE Kernel and Convolutional Neural Networks, ICC 2024-IEEE International Conference on Communications, IEEE, 2024, pp. 5347-5352.

[70] J. Xiang, E. Shlizerman, Tkil: Tangent kernel optimization for class balanced incremental learning, Proceedings of the IEEE/CVF International Conference on Computer Vision, 2023, pp. 3529-3539.

[71] L. Ma, Y. Qiao, R. Wang, H. Chen, G. Liu, H. Xiao, R. Dai, Machine Learning Models Decoding the Association Between Urinary Stone Diseases and Metabolic Urinary Profiles, Metabolites 14(12) (2024) 674.

[72] H. Tu, Y. Shi, M. Xu, Integrating conditional shape embedding with generative adversarial network-to assess raster format architectural sketch, 2023 Annual Modeling and Simulation Conference (ANNSIM), IEEE, 2023, pp. 560-571.

[73] Y. Jin, M. Choi, G. Verma, J. Wang, S. Kumar, Mm-soc: Benchmarking multimodal large language models in social media platforms, arXiv preprint arXiv:2402.14154 (2024).

[74] Z. Gao, Y. Wu, X. Bu, T. Yu, J. Yuan, Y. Jia, Learning a robust representation via a deep network on symmetric positive definite manifolds, Pattern Recognition 92 (2019) 1-12.

[75] J. Zhang, X. Wang, Y. Jin, C. Chen, X. Zhang, K. Liu, Prototypical Reward Network for Data-Efficient RLHF, arXiv preprint arXiv:2406.06606 (2024).

[76] K. Xu, L. Chen, S. Wang, Are KAN Effective for Identifying and Tracking Concept Drift in Time Series?, arXiv preprint arXiv:2410.10041 (2024).

[77] Z. Ke, J. Xu, Z. Zhang, Y. Cheng, W. Wu, A Consolidated Volatility Prediction with Back Propagation Neural Network and Genetic Algorithm, arXiv preprint arXiv:2412.07223 (2024).

[78] K. Xu, L. Chen, S. Wang, Data-driven kernel subspace clustering with local manifold preservation, 2022 IEEE International Conference on Data Mining Workshops (ICDMW), IEEE, 2022, pp. 876-884.

[79] J. Zhang, X. Wang, W. Ren, L. Jiang, D. Wang, K. Liu, RATT: AThought Structure for Coherent and Correct LLMReasoning, arXiv preprint arXiv:2406.02746 (2024).

[80] T. Tian, S. Jia, J. Lin, Z. Huang, K.O. Wang, Y. Tang, Enhancing Industrial Management through AI Integration: A Comprehensive Review of Risk Assessment, Machine Learning Applications, and Data-Driven Strategies, Economics & Management Information (2024) 1-18.

[81] P. Zhao, L. Lai, Minimax Optimal Q Learning with Nearest Neighbors, IEEE Transactions on Information Theory (2024).

[82] Z. Ding, J. Tian, Z. Wang, J. Zhao, S. Li, Data imputation using large language model to accelerate recommendation system, arXiv preprint arXiv:2407.10078 (2024).

[83] T. Zheng, Y. Jin, H. Zhao, Z. Ma, Y. Chen, K. Xu, Learning coverage paths in unknown environments with deep reinforcement learning.

[84] B. Dang, W. Zhao, Y. Li, D. Ma, Q. Yu, E.Y. Zhu, Real-Time pill identification for the visually impaired using deep learning, arXiv preprint arXiv:2405.05983 (2024).

[85] J. Liang, X. Gu, Y. Chen, F. Ni, T. Zhang, A novel pavement mean texture depth evaluation strategy based on three-dimensional pavement data filtered by a new filtering approach, Measurement 166 (2020) 108265.

[86] K. Li, J. Wang, X. Wu, X. Peng, R. Chang, X. Deng, Y. Kang, Y. Yang, F. Ni, B. Hong, Optimizing automated picking systems in warehouse robots using machine learning, arXiv preprint arXiv:2408.16633 (2024).

[87] X. Fang, S. Si, G. Sun, Q.Z. Sheng, W. Wu, K. Wang, H. Lv, Selecting workers wisely for crowdsourcing when copiers and domain experts co-exist, Future Internet 14(2) (2022) 37.

[88] X. Liu, Z. Dong, P. Zhang, Tackling data bias in music-avqa: Crafting a balanced dataset for unbiased question-answering, Proceedings of the IEEE/CVF Winter Conference on Applications of Computer Vision, 2024, pp. 4478-4487.

[89] K.H. Wong, J. Cao, Y. Yang, W. Li, J. Wang, Z. Yao, S. Xu, E.A.C. Ku, C.O. Wong, D. Leung, BigARM: A big-data-driven airport resource management engine and application

tools, Database Systems for Advanced Applications: 25th International Conference, DASFAA 2020, Jeju, South Korea, September 24–27, 2020, Proceedings, Part III 25, Springer, 2020, pp. 741-744.

[90] J. Tang, W. Du, B. Wang, W. Zhou, S. Mei, T. Xue, X. Xu, H. Zhang, Character recognition competition for street view shop signs, National Science Review 10(6) (2023) nwad141.

[91] K. Li, J. Chen, D. Yu, T. Dajun, X. Qiu, L. Jieting, S. Baiwei, Z. Shengyuan, Z. Wan, R. Ji, Deep reinforcement learning-based obstacle avoidance for robot movement in warehouse environments, arXiv preprint arXiv:2409.14972 (2024).

[92] Z. Lin, A. Ruszczyński, An Integrated Transportation Distance between Kernels and Approximate Dynamic Risk Evaluation in Markov Systems, SIAM Journal on Control and Optimization 61(6) (2023) 3559-3583.

[93] Y. Tao, Y. Jia, N. Wang, H. Wang, The fact: Taming latent factor models for explainability with factorization trees, Proceedings of the 42nd international ACM SIGIR conference on research and development in information retrieval, 2019, pp. 295-304.

[94] D. Liu, R. Waleffe, M. Jiang, S. Venkataraman, Graphsnapshot: Graph machine learning acceleration with fast storage and retrieval, arXiv preprint arXiv:2406.17918 (2024).

[95] Z. Lin, A. Ruszczyński, Stochastic Kernel Approximation by Transportation Distance Method, 2024 IISE Annual Conference and Expo, IISE, 2024.

[96] D. Liu, MT2ST: Adaptive Multi-Task to Single-Task Learning, arXiv preprint arXiv:2406.18038 (2024).

[97] Z.-h. Li, J.-x. Cui, H.-p. Lu, F. Zhou, Y.-l. Diao, Z.-x. Li, Prediction model of measurement errors in current transformers based on deep learning, Review of Scientific Instruments 95(4) (2024).

[98] J. Tian, J. Zhao, Z. Wang, Z. Ding, Mmrec: Llm based multi-modal recommender system, arXiv preprint arXiv:2408.04211 (2024).

[99] Z. Li, J. Cao, Z. Yao, W. Li, Y. Yang, J. Wang, Recursive Balanced k-Subset Sum Partition for Rule-constrained Resource Allocation, Proceedings of the 29th ACM

International Conference on Information & Knowledge Management, 2020, pp. 2121-2124.

[100] F. Guo, H. Mo, J. Wu, L. Pan, H. Zhou, Z. Zhang, L. Li, F. Huang, A hybrid stacking model for enhanced short-term load forecasting, Electronics 13(14) (2024) 2719.

[101] Z. Dong, P. Polak, CP-PINNs: Data-Driven Changepoints Detection in PDEs Using Online Optimized Physics-Informed Neural Networks, 2024 Conference on AI, Science, Engineering, and Technology (AIxSET), IEEE, 2024, pp. 90-97.

[102] Z. Xiao, Z. Mai, Z. Xu, Y. Cui, J. Li, Corporate Event Predictions Using Large Language Models, 2023 10th International Conference on Soft Computing & Machine Intelligence (ISCMI), IEEE, 2023, pp. 193-197.

[103] Z. Ke, Y. Yin, Tail Risk Alert Based on Conditional Autoregressive VaR by Regression Quantiles and Machine Learning Algorithms, arXiv preprint arXiv:2412.06193 (2024).

[104] Y. Yan, F. Guo, H. Mo, X. Huang, Hierarchical Tracking Control for a Composite Mobile Robot Considering System Uncertainties, 2024 16th International Conference on Computer and Automation Engineering (ICCAE), IEEE, 2024, pp. 512-517.

[105] D. Zhang, D. Chen, P. Zhi, Y. Chen, Z. Yuan, C. Li, R. Zhou, Q. Zhou, MapExpert: Online HD Map Construction with Simple and Efficient Sparse Map Element Expert, arXiv preprint arXiv:2412.12704 (2024).

[106] L. Jiang, X. Yang, C. Yu, Z. Wu, Y. Wang, Advanced AI framework for enhanced detection and assessment of abdominal trauma: Integrating 3D segmentation with 2D CNN and RNN models, 2024 3rd International Conference on Robotics, Artificial Intelligence and Intelligent Control (RAIIC), IEEE, 2024, pp. 337-340.

[107] Y. Tao, Meta Learning Enabled Adversarial Defense, 2023 IEEE International Conference on Sensors, Electronics and Computer Engineering (ICSECE), IEEE, 2023, pp. 1326-1330.

[108] Z. Fu, K. Wang, W. Xin, L. Zhou, S. Chen, Y. Ge, D. Janies, D. Zhang, Detecting Misinformation in Multimedia Content through Cross-Modal Entity Consistency: A Dual Learning Approach, (2024).

[109] J. Wang, J. Cao, S. Wang, Z. Yao, W. Li, IRDA: Incremental reinforcement learning for dynamic resource allocation, IEEE Transactions on Big Data 8(3) (2020) 770-783.

[110] Y. Shi, A. Economou, Dougong Revisited: A Parametric Specification of Chinese Bracket Design in Shape Machine, International Conference on-Design Computing and Cognition, Springer, 2024, pp. 233-249.

[111] Z. Xiao, Z. Mai, Y. Cui, Z. Xu, J. Li, Short Interest Trend Prediction with Large Language Models, Proceedings of the 2024 International Conference on Innovation in Artificial Intelligence, 2024, pp. 1-1.

[112] Y. Chen, Q. Fu, G. Fan, L. Du, J.-G. Lou, S. Han, D. Zhang, Z. Li, Y. Xiao, Hadamard adapter: An extreme parameter-efficient adapter tuning method for pre-trained language models, Proceedings of the 32nd ACM International Conference on Information and Knowledge Management, 2023, pp. 276-285.

[113] J. Xiang, S. Colburn, A. Majumdar, E. Shlizerman, Knowledge distillation circumvents nonlinearity for optical convolutional neural networks, Applied Optics 61(9) (2022) 2173-2183.

[114] X. Fang, S. Si, G. Sun, W. Wu, K. Wang, H. Lv, A Domain-Aware Crowdsourcing System with Copier Removal, International Conference on Internet of Things, Communication and Intelligent Technology, Springer, 2022, pp. 761-773.